# Shallow Univariate ReLu Networks as Splines: Initialization, Loss Surface, Hessian, & Gradient Flow Dynamics

**Justin Sahs**[1*] **Ryan Pyle**[1*] **Aneel Damaraju** [2]
**Josue Ortega Caro**[1] **Onur Tavaslioglu**[2] **Andy Lu**[1]
**Ankit Patel**[1,2]
[1] Baylor College of Medicine
[2] Rice University
{Justin.Sahs, Ryan.Pyle, Josue.OrtegaCaro, Ankit.Patel}@bcm.edu
{amd18, hl61}@rice.edu, onurtav@amazon.com

## Abstract

Understanding the learning dynamics and inductive bias of neural networks (NNs) is hindered by the opacity of the relationship between NN parameters and the function represented. We propose reparametrizing ReLU NNs as continuous piecewise linear splines. Using this spline lens, we study learning dynamics in shallow univariate ReLU NNs, finding unexpected insights and explanations for several perplexing phenomena. We develop a surprisingly simple and transparent view of the structure of the loss surface, including its critical and fixed points, Hessian, and Hessian spectrum. We also show that standard weight initializations yield very flat functions, and that this flatness, together with overparametrization and the initial weight scale, is responsible for the strength and type of implicit regularization, consistent with recent work [23]. Our implicit regularization results are complementary to recent work [22], done independently, which showed that initialization scale critically controls implicit regularization via a kernel-based argument. Our spline-based approach reproduces their key implicit regularization results but in a far more intuitive and transparent manner. Going forward, our spline-based approach is likely to extend naturally to the multivariate and deep settings, and will play a foundational role in efforts to understand neural networks. Videos of learning dynamics using a spline-based visualization are available at `http://shorturl.at/tFWZ2`.

## 1 Introduction

Despite widespread use of deep learning, theoretical understanding of some fundamental properties, e.g. the root cause of implicit regularization and generalization achieved by overparameterized networks, remains poorly understood [24]. In contrast to other machine learning techniques, continuing to add parameters to a deep network (beyond zero training loss) tends to *improve* generalization performance. This has even been observed for networks that are massively overparameterized, wherein, according to traditional ML theory, they should (over)fit the training data [17]. How does training networks with excess capacity lead to generalization? And how can generalization error decrease with overparameterization?

Here we show that the spline view allows us a new, useful lens with which to explore and understand these issues. In particular, we focus on shallow fully connected univariate ReLU networks, whose parameters will always result in a Continuous Piecewise Linear (CPWL) output. We provide theoretical results for shallow networks, with experiments confirming these qualitative results.

Our approach is related to previous work [20, 4, 8] in that we wish to characterize parameterization and generalization. We differ from these other works by focusing on small width networks, rather than massively overparametrized or infinite width networks, and by using a spline parameterization to study properties such as smoothness of the approximated function. Previous work [2] has hinted at the importance of small norm initialization, but the spline perspective allows us to prove implicit regularization properties in shallow networks.

Concurrent with this work, [22] proved a number of results similar to those in Section 2.5. Overall, we feel that our results provide clear geometric/functional interpretations that are missing in [22]. We include a detailed comparison at the end of that section.

**Main Contributions** The main contribution of this work are as follows:

- *Initialization: Increasingly Flat with Width.* In the spline perspective, neural network parameters determine the locations of breakpoints and their delta-slopes (defined in Section 2.1) in the CPWL reparameterization. We prove that, for common initializations, these distributions are mean 0 with low standard deviation. Notably, the delta-slope distribution becomes increasingly concentrated near 0 as the width of the network increases, leading to flatter initial approximations.
- *A characterization of the Loss Surface, Critical Points and Hessian, revealing that Flat Minima are due to degeneracy of breakpoints.* We fully characterize the Hessian of the loss surface at critical points, revealing that the Hessian is positive semi-definite, with 0 eigenvalues occurring when elements of its Gram matrix set are linearly dependent. We fully characterize the ways this can occur, including whenever multiple breakpoints share the same active data - thus we expect that at any given critical point, the loss surface will be flat in many directions.
- *Implicit Regularization is due to Flat Initialization in the Overparameterized Regime.* We find that implicit regularization in overparametrized FC ReLU nets is due to three factors: (i) the very flat initialization, (ii) the curvature-based parametrization of the approximating function (breakpoints and delta-slopes) and (iii) the role of gradient descent (GD) in preserving initialization and regularizing via curvature. In particular, the global, rather than local, impact of breakpoints and delta-slopes regularizes the delta-slopes, in effect ‘distributing’ the work over units. All else equal, these nonlocal effects mean that more overparameterization leads to each unit mattering less, and thus typically resulting in better generalization due to implicit regularization [18, 17].
- *The Type of Implicit Regularization is controlled by the Initialization Weight Scale $\alpha$.* The spline perspective makes it clear that the weight initialization scale $\alpha$, which has no effect on the function, but does affect gradients, determines the different kinds of implicit regularization (e.g. kernel and rich regimes).

# 2 Theoretical Results

## 2.1 ReLU Nets as Splines: From Weights to Breakpoints & Delta-Slopes

Consider a fully connected ReLU neural net $\hat{f}_\theta(x)$ with a single hidden layer of width $H$, scalar input $x \in \mathbb{R}$ and scalar output $y \in \mathbb{R}$. $\hat{f}(\cdot;\theta)$ is continuous piecewise linear function (CPWL) since the ReLU nonlinearity is CPWL. We want to understand the *function* implemented by this neural net, and so we ask: How do the CPWL parameters relate to the NN parameters? We answer this by transforming from the NN parameterization of weights and biases to two CPWL *spline* parameterizations:

$$\begin{aligned}\hat{f}(x;\theta_{\text{NN}}) &\triangleq \sum_{i=1}^{H} v_i(w_i x + b_i)_+ + b_0 \\ &= b_0 + \sum_{i=1}^{H} \mu_i (x-\beta_i)_{s_i} \triangleq \hat{f}(x;\theta_{\text{BDSO}}) \qquad (1) \\ &= b_0 + \sum_{p=1}^{P} [\![\beta_p \le x < \beta_{p+1}]\!]\,(m_p x + \gamma_p) \triangleq \hat{f}(x;\theta_{\text{PWL}}) \qquad (2)\end{aligned}$$

Here the NN parameters $\theta_{\text{NN}} \triangleq \{b_0\} \cup \{(w_i, b_i, v_i)\}_{i=1}^H$ denote the input weight, bias, and output weight respectively of neuron $i$, and $(\cdot)_+ \triangleq \max\{0, \cdot\}$ denotes the ReLU function. The Iversen bracket $[\![b]\!]$ is 1 when the condition $b$ is true, and 0 otherwise, and we let

$$(x - \beta_i)_{s_i} \triangleq (x - \beta_i) \begin{cases} [\![x > \beta_i]\!], & s_i = 1 \\ [\![x < \beta_i]\!], & s_i = -1 \end{cases}$$

The first CPWL spline parametrization is the ***B**reakpoint, **D**elta-**S**lope, **O**rientation* (BDSO) parametrization $\theta_{\text{BDSO}} \triangleq \{b_0\} \cup \{(\beta_i, \mu_i, s_i)\}_{i=1}^H$, where $\beta_i \triangleq -\frac{b_i}{w_i}$ is (the x-coordinate of) the *breakpoint* (or *knot*) induced by neuron $i$, $\mu_i \triangleq w_i v_i$ is the *delta-slope* contribution of neuron $i$, and $s_i \triangleq \operatorname{sgn} w_i \in \{\pm 1\}$ is the *orientation* of $\beta_i$ (left for $s_i = -1$, right for $s_i = +1$).

Intuitively, in a good fit the breakpoints $\beta_i$ will congregate in areas of high curvature in the ground truth function $|f''(x)| \geq 0$, while delta-slopes $\mu_i$ will actually implement the needed curvature by changing the slope by $\mu_i$ from one piece $p(i)$ to the next $p(i) + 1$. As the number of pieces grows, the linear spline approximation will improve, and the delta-slopes (scaled by the piece lengths) approach the true curvature of $f$: $\lim_{P\to\infty} \mu_{p(i)}/(\beta_p - \beta_{p-1}) \to f''(x = \beta_i)$.

The second parametrization is the canonical one for PWL functions: $\theta_{\text{PWL}} \triangleq \{b_0\} \cup \{(\beta_p, m_p, \gamma_p)\}_{p=1}^P$, where $\beta_0 < \beta_1 < \ldots < \beta_p \triangleq -\frac{b_{p(i)}}{w_{p(i)}} < \ldots < \beta_P$ is the sorted list of the $x$-coordinates of the $P \triangleq H + 1$ breakpoints (or knots), $m_p, \gamma_p$ are the slope and y-intercept of piece $p$. Note that the breakpoints induce two partitions: (i) a partition of the *domain* $\mathcal{X} = \cup_{p=0}^P \mathcal{X}_p \triangleq \cup_{p=0}^P [\beta_p, \beta_{p+1})$; and (ii) a partition of the *training data* $\mathcal{D}_x \triangleq \{x_n\}_{n=1}^N = \cup_{p=0}^P \pi_p \triangleq \cup_{p=0}^P [\beta_p, \beta_{p+1}) \cap \mathcal{D}_x$, where we use the convention that $\beta_0 \triangleq x_{\min}$ and $\beta_{P+1} \triangleq x_{\max}$.

**Degeneracy in ReLu NN Parametrization.** An important observation is that the ReLu NN parametrization is redundant: for every function $\hat{f}$ represented by Equation (1) there exists infinitely many transformations of the parameters $\theta'_{\text{NN}} \equiv \mathcal{R}(\theta_{\text{NN}})$ s.t. the transformed function $\hat{f}(x; \theta'_{\text{NN}}) = \hat{f}(x; \theta_{\text{NN}})$ i.e. the function is invariant. Mathematically, such invariant transformations $\mathcal{R}$ include (i) permutations of the hidden units and (ii) scalings of the weights and biases of the form $w_i \mapsto \alpha_i w_i, b_i \mapsto \alpha_i b_i, v_i \mapsto \alpha_i^{-1} v_i$ for $\alpha_i \in \mathbb{R}_{>0}$. These transformations exhaust the list of invariant ones [19]. The set $\mathcal{G}$ of such function-invariant transformations together with function composition $\circ$ forms a group. Unlike the ReLu NN parametrization, the CPWL spline parametrization is unique due to the sorting of the breakpoints and direct representation of the slopes per piece.

We note that the BDSO parametrization of a ReLU NN is closely related to but different than a traditional roughness-minimizing $m$-th order spline parametrization $\hat{f}_{\text{spline}}(x) \triangleq \sum_{i=1}^K \mu_i (x - \beta_i)_+^m + \sum_{j=0}^m c_j x^j$. BDSO (i) lacks the base polynomial, and (ii) has two possible breakpoint orientations $s_i \in \{\pm 1\}$ whereas the spline only has one. We note in passing that adding in the base polynomial (for linear case $m = 1$) into the BDSO parametrization yields a ReLU ResNet parametrization. We believe this is a novel viewpoint that may shed more light on the origin of the effectiveness of ResNets, but we leave it for future work.

Considering these parameterizations (especially the BDSO parameterization) provides a new, useful lens with which to analyze neural nets, enabling us to reason more easily and transparently about the initialization, loss surface, and training dynamics. The benefits of this approach derive from two main properties: (1) that we have 'modded out' the degeneracies in the NN parameterization and (2) the loss depends on the NN parameters $\theta_{\text{NN}}$ only through the BDSO parameters (the approximating function) $\theta_{\text{BDSO}}$ i.e. $\ell(\theta_{\text{NN}}) = \ell(\theta_{\text{BDSO}}(\theta_{\text{NN}}))$, analogous to the concept of a minimum sufficient statistic in exponential family models. Much recent related work has also veered in this direction, analyzing function space [11, 6].

### 2.2 Random Initialization in Function Space

We now study the random initializations commonly used in deep learning in function space. These include the independent Gaussian initialization, with $b_i \sim \mathcal{N}(0, \sigma_b)$, $w_i \sim \mathcal{N}(0, \sigma_w)$, $v_i \sim \mathcal{N}(0, \sigma_v)$, and independent Uniform initialization, with $b_i \sim \mathcal{U}[-a_b, a_b]$, $w_i \sim \mathcal{U}[-a_w, a_w]$, $v_i \sim \mathcal{U}[-a_v, a_v]$. We find that common initializations result in flat functions, becoming flatter with increasing width.

**Theorem 1.** *Consider a fully connected ReLU neural net with scalar input and output, and a single hidden layer of width $H$. Let the weights and biases be initialized randomly according to a zero-mean Gaussian or Uniform distribution. Then the induced distributions of the function space parameters (breakpoints $\beta$, delta-slopes $\mu$) are as follows:*

*(a) Under an independent Gaussian initialization,*

$$p_{\beta,\mu}(\beta_i, \mu_i) = \frac{1}{2\pi\sigma_v\sqrt{\sigma_b^2+\sigma_w^2\beta_i^2}} \exp\left[-\frac{|\mu_i|\sqrt{\sigma_b^2+\sigma_w^2\beta_i^2}}{\sigma_b\sigma_v\sigma_w}\right]$$

*(b) Under an independent Uniform initialization,*

$$p_{\beta,\mu}(\beta_i, \mu_i) = \frac{[\![|\mu_i| \le \min\{\frac{a_b a_v}{|\beta_i|}, a_w, a_v\}]\!]}{4a_b a_w a_v} \times \left(\min\{\frac{a_b}{|\beta_i|}, a_w\} - \frac{|\mu_i|}{a_v}\right)$$

Using this result, we can immediately derive marginal and conditional distributions for the breakpoints and delta-slopes.

**Corollary 1.** *Consider the same setting as Theorem 1.*

*(a) In the case of an independent Gaussian initialization,*

$$p_\beta(\beta_i) = \mathrm{Cauchy}\left(\beta_i; 0, \frac{\sigma_b}{\sigma_w}\right)$$
$$p_\mu(\mu_i) = \frac{G_{0,2}^{2,0}\left(\frac{\mu_i^2}{4\sigma_v^2\sigma_w^2}\middle| 0, 0\right)}{2\pi\sigma_v\sigma_w} = \frac{K_0\left(\frac{|\mu_i|}{\sigma_v\sigma_w}\right)}{\pi\sigma_v\sigma_w}$$
$$p_{\mu|\beta}(\mu_i|\beta_i) = \mathrm{Laplace}\left(\mu_i; 0, \frac{\sigma_b\sigma_v\sigma_w}{\sqrt{\sigma_b^2+\sigma_w^2\beta_i^2}}\right)$$

*where $G_{pq}^{nm}(\cdot|\cdot)$ is the Meijer G-function and $K_\nu(\cdot)$ is the modified Bessel function of the second kind.*

*(b) In the case of an independent Uniform initialization,*

$$p_\beta(\beta_i) = \frac{1}{4a_b a_w}\left(\min\left\{\frac{a_b}{|\beta_i|}, a_w\right\}\right)^2$$
$$p_\mu(\mu_i) = \frac{[\![-a_w a_v \le \mu_i \le a_w a_v]\!]}{2a_w a_v} \log\frac{a_w a_v}{|\mu_i|}$$
$$p_{\mu|\beta}(\mu_i|\beta_i) = \mathrm{Tri}(\mu_i; a_v \min\{a_b/|\beta_i|, a_w\})$$

*where $\mathrm{Tri}(\cdot; a)$ is the symmetric triangular distribution with base $[-a, a]$ and mode $0$.*

**Implications.** Corollary 1 implies that the breakpoint density drops quickly away from the origin for common initializations. If $f$ has significant curvature far from the origin, then it may be far more difficult to fit. We show that this is indeed the case by training a shallow ReLU NN with an initialization that does not match the underlying curvature, with training becoming easier if the initial breakpoint distribution better matches the function curvature. We also show that during training, breakpoint distributions move to better match the underlying function curvature, and that this effect increases with depth (see Section 3 and Table 1). This implies that a data-dependent initialization, with more breakpoints near areas of high curvature, could potentially be faster and easier to train.

Next, we consider the typical Gaussian He [12] or Glorot [9] initializations. In the He initialization, we have $\sigma_w = \sqrt{2}$, $\sigma_v = \sqrt{2/H}$. In the Glorot initalization, we have $\sigma_w = \sigma_v = \sqrt{2/(H+1)}$. We wish to consider their effect on the smoothness of the initial function approximation. From here on, we measure the smoothness using a roughness metric, defined as $\rho \triangleq \int |\hat{f}''(x; \theta_{\mathrm{BDSO}})|^2 \, \mathrm{d}x = \sum_i \mu_i^2$, where lower roughness indicates a smoother approximation.

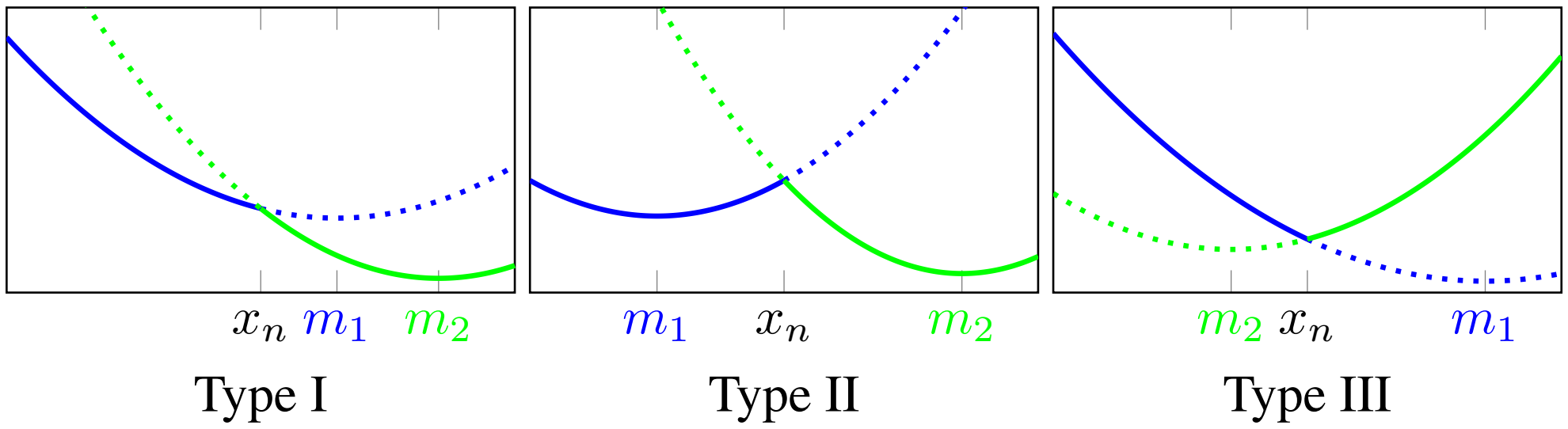

Figure 1: Classification of Critical Points

**Theorem 2.** *Consider the initial roughness $\rho_0$ under a Gaussian initialization. In the He initialization, we have that the tail probability is given by*

$$\mathbb{P}[\rho_0 - \mathbb{E}[\rho_0] \geq \lambda] \leq \frac{1}{1 + \frac{\lambda^2 H}{128}},$$

*where $\mathbb{E}[\rho_0] = 4$. In the Glorot initialization, we have that the tail probability is given by*

$$\mathbb{P}[\rho_0 - \mathbb{E}[\rho_0] \geq \lambda] \leq \frac{1}{1 + \frac{\lambda^2 (H+1)^4}{128H}},$$

*where $\mathbb{E}[\rho_0] = \frac{4H}{(H+1)^2} = O\left(\frac{1}{H}\right)$.*

Thus, as the width $H$ increases, the distribution of the roughness of the initial function $\hat{f}_0$ gets tighter around its mean. In the case of the He initialization, this mean is constant; in the Glorot initialization, it decreases with $H$. In either case, for reasonable widths, the initial roughness is small with high probability. This smoothness has implications for the implicit regularization/generalization phenomenon observed in recent work [18]. Finally, as breakpoints are necessary to fit curvature, the final learned function and its generalization could critically depend on the initial distribution. We cover this idea in greater depth in Section 2.5, with experimental results in Table 1.

*Related Work.* Several recent works analyze the random initialization in deep networks. However, there are two main differences, First, they focus on the infinite width case [20, 13, 14, 16] and can thus use the Central Limit Theorem (CLT), whereas we focus on finite width case and cannot use the CLT, thus requiring nontrivial mathematical machinery (see Supplement for detailed proofs). Second, they focus on the activations as a function of input whereas we also compute the joint densities of the BDSO parameters i.e. breakpoints and delta-slopes. The latter is particularly important for understanding the non-uniform density of breakpoints away from the origin as noted above.

### 2.3 Loss Surface in the Spline Parametrization

We now consider the squared loss $\frac{1}{2}\sum_{n=1}^{N}(\hat{f}(x_n;\theta) - y_n)^2$ as a function of either the NN parameters $\ell(\theta_{\text{NN}})$ or the BDSO parameters $\tilde{\ell}(\theta_{\text{BDSO}})$.

**Theorem 3.** *The loss function $\tilde{\ell}(\theta_{BDSO} = (\boldsymbol{\beta}, \boldsymbol{\mu}), \mathbf{s})$ is a continuous piecewise quadratic spline. Furthermore, the 1-dimensional slice $\tilde{\ell}(\beta_i; \boldsymbol{\beta}_{-i}, \boldsymbol{\mu}, \mathbf{s})$ is also a continuous piecewise quadratic spline in $\beta_i$ with knots at datapoints $\{x_n\}_{n=1}^{N}$. Let $p_1(\beta_i)$ (resp. $p_2(\beta_i)$) be the quadratic function equal to $\tilde{\ell}(\beta_i; \boldsymbol{\beta}_{-i}, \boldsymbol{\mu}, \mathbf{s})$ for $\beta_i \in [x_{n-1}, x_n]$ (resp. $[x_n, x_{n+1}]$), which both have positive curvature, and let $m_j \triangleq \arg\min p_j(\beta_i)$. Then, with measure 1, the knots $x_n$ fall into one of three types as shown in Figure 1: (Type I) $m_1, m_2 < x_n$, or $x_n < m_1, m_2$, (Type II) $m_1 < x_n < m_2$ (Type III) $m_2 < x_n < m_1$.*

Our Hessian analysis reveals that the loss surface is continuous piece-wise quadratic (CPWQ), where the piece boundaries correspond to changing activation patterns as a breakpoint moves past a datapoint. Thus, for $\beta_i$ in an $\epsilon$ ball around datapoint $x_n$, the loss surface $\ell(\beta_i, \theta)$ for $\theta$ fixed consists of two

$$
\mathbf{H}_\ell(\theta) = \begin{pmatrix} \ddots & & & & \\ \langle v_j \mathbf{x}_j, v_i \mathbf{x}_i \rangle & \langle v_j \mathbf{x}_j, w_i \mathbf{x}_i + b_i \mathbf{1}_i \rangle & \langle v_j \mathbf{x}_j, v_i \mathbf{1}_i \rangle & \cdots & \langle v_j \mathbf{x}_j, \mathbf{1} \rangle \\ \langle w_j \mathbf{x}_j + b_j \mathbf{1}_j, v_i \mathbf{x}_i \rangle & \langle w_j \mathbf{x}_j + b_j \mathbf{1}_j, w_i \mathbf{x}_i + b_i \mathbf{1}_i \rangle & \langle w_j \mathbf{x}_j + b_j \mathbf{1}_j, v_i \mathbf{1}_i \rangle & \cdots & \langle w_j \mathbf{x}_j + b_j \mathbf{1}_j, \mathbf{1} \rangle \\ \langle v_j \mathbf{1}_j, v_i \mathbf{x}_i \rangle & \langle v_j \mathbf{1}_j, w_i \mathbf{x}_i + b_i \mathbf{1}_i \rangle & \langle v_j \mathbf{1}_j, v_i \mathbf{1}_i \rangle & \cdots & \langle v_j \mathbf{1}_j, \mathbf{1} \rangle \\ \vdots & & & \ddots & \vdots \\ \langle \mathbf{1}, v_i \mathbf{x}_i \rangle & \langle \mathbf{1}, w_i \mathbf{x}_i + b_i \mathbf{1}_i \rangle & \langle \mathbf{1}, v_i \mathbf{1}_i \rangle & \cdots & \langle \mathbf{1}, \mathbf{1} \rangle \end{pmatrix} \tag{3}
$$

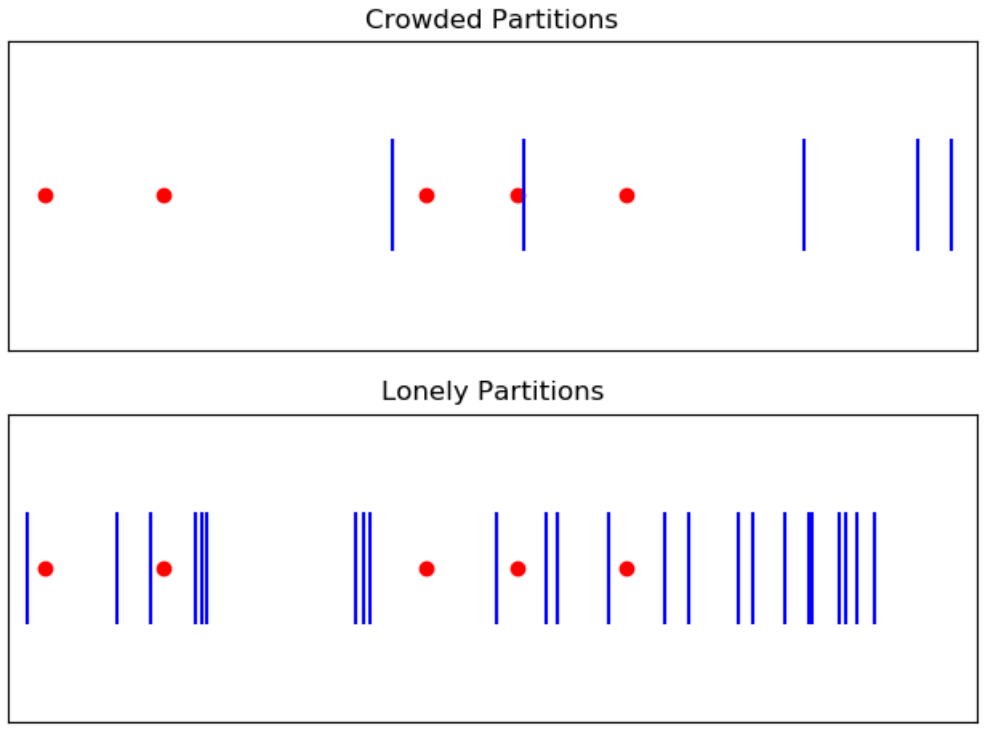

Figure 2: Breakpoints (blue bars) vs datapoints (red points). A lonely partition is when a datapoint is isolated - overparameterization makes this increasingly likely

connected quadratics joined at $\beta_i = x_n$. We can fully characterize this intersection as one of three types depending on the relationship between the minima of these parabolas (recall Theorem 3). In Case (i), both minima are on the same side of the knot $x_n$, and $\beta_i$ will freely travel past $x_n$ towards the minima. In Case (ii), the minimum of each parabola is located on its active side with respect to the knot $x_n$, making $x_n$ a repeller for $\beta_i$. In Case (iii), the minimum of each parabola is located on the inactive side of $x_n$, making it an attractor for $\beta_i$. This classification of the knots will be useful for understanding learning dynamics in Section 2.5.

**Theorem 4.** *Consider some arbitrary $\theta^*_{NN}$ such that there exists at least one neuron that is active on all data. Then, $\theta^*_{NN}$ is a critical point of $\tilde{\ell}(\cdot)$ if and only if for all pieces $p \in [P]$, $\hat{f}(\cdot; \theta^*_{NN})|_{\pi_p} = m_p x + \gamma_p$ is an Ordinary Least Squares (OLS) fit of the training data contained in piece $p$.*

An open question is how many such critical points exist. A starting point is to consider that there are $C(N+H, H) \triangleq (N+H)!/N!H!$ possible partitions of the data. Not every such partition will admit a piecewise-OLS solution which is also continuous, and it is difficult to analytically characterize such solutions, so we resort to simulation and find a lower bound that suggests the number of critical points grows at least polynomially in $N$ and $H$ (Figure 7).

Using Theorem 4, we can characterize growth of global minima in the overparameterized case. Call a partition $\Pi$ *lonely* if each piece $\pi_p \in \Pi$ contains at most one datapoint. Then, we can prove the following results:

**Lemma 1.** *For any lonely partition $\Pi$, there are infinitely many parameter settings $\theta_{BDSO}$ that induce $\Pi$ and are global minima with $\tilde{\ell}(\theta_{BDSO}) = 0$. Furthermore, in the overparametrized regime $H \geq cN$ for some constant $c \geq 1$, the total number of lonely partitions, and thus a lower bound on the total number of* global *minima of $\tilde{\ell}$ is $\binom{H+1}{N} = O(N^{cN})$.*

**Remark 1.** *Suppose that the $H$ breakpoints are uniformly spaced and that the $N$ datapoints are uniformly distributed within the region of breakpoints. Then in the overparametrized regime $H \geq dN^2$ for some constant $d \geq 1$, the induced partition $\Pi$ is lonely with high probabilility $1 - e^{-N^2/(H+1)} = 1 - e^{-1/d}$.*

Thus, with only $O(N^2)$ hidden units, we can almost guarantee a lonely partition at initialization. However, this makes optimization easier but is not sufficient to guarantee that learning will converge to a global minima, as breakpoint dynamics could change the partition to crowded (see Section 2.4). Note how simple and transparent the spline-based explanation is for why overparametrization makes optimization easier. For brevity, we include experiments testing our theory of loneliness in Figure 11 in the Supplement.

**The Gradient and Hessian of the Loss $\ell(\theta_{\text{NN}})$** Let $\hat{\boldsymbol{\epsilon}} \in \mathbb{R}^{N\times 1}$ denote the vector of residuals $\mathbf{y} - \hat{\mathbf{f}}$, i.e. $\hat{\epsilon}_n = f(x_n) - \hat{f}(x_n)$, let $\mathbf{1}_i \in \mathbb{R}^{N\times 1}$ denote the activation pattern of neuron $i$ over all $N$ inputs, $\mathbf{1}_{i,n} = [\![w_i x_n + b_i > 0]\!]$, and let $\hat{\boldsymbol{\epsilon}}_i \triangleq \hat{\boldsymbol{\epsilon}} \odot \mathbf{1}_i \in \mathbb{R}^{N\times 1}$ and $\mathbf{x}_i \triangleq \mathbf{x} \odot \mathbf{1}_i \in \mathbb{R}^{N\times 1}$ denote the *relevant* residuals and inputs for neuron $i$, respectively. Using this notation, the gradient of $\ell(\theta_{\text{NN}})$ can be expressed as

$$\begin{aligned} \frac{\partial \ell}{\partial b_0} &= -\langle \hat{\boldsymbol{\epsilon}}, \mathbf{1}\rangle & \frac{\partial \ell}{\partial w_i} &= -\langle \hat{\boldsymbol{\epsilon}}_i, v_i \mathbf{x}_i\rangle \\ \frac{\partial \ell}{\partial v_i} &= -\langle \hat{\boldsymbol{\epsilon}}_i, w_i \mathbf{x}_i + b_i \mathbf{1}_i\rangle & \frac{\partial \ell}{\partial b_i} &= -\langle \hat{\boldsymbol{\epsilon}}_i, v_i \mathbf{1}_i\rangle \end{aligned} \tag{4}$$

From this, we can derive expressions for the Hessian of the loss $\ell(\theta_{\text{NN}})$ at any parameter $\theta_{\text{NN}}$.

**Theorem 5.** *Let $\theta^*$ be a critical point of $\ell(\cdot)$ such that $\beta_i^*(\theta^*) \neq x_n$ for all $i \in [H]$ and for all $n \in [N]$. Then the Hessian $\mathbf{H}_\ell(\theta^*)$ is the positive semi-definite Gram matrix of the set of $3H+1$ vectors*

$$\mathcal{B} \triangleq \{v_i \mathbf{x}_i, w_i \mathbf{x}_i + b_i \mathbf{1}_i, v_i \mathbf{1}_i\}_{i=1}^H \cup \{\mathbf{1}\},$$

*as shown in Equation* (3). *Thus, $\mathbf{H}_\ell(\theta^*)$ is positive definite iff the vectors of this set are linearly independent.*

**Corollary 2.** *Consider the setting of Theorem 5. There are seven cases where linear independence fails, of which the most important are (i) when distinct neurons share activation patterns, and (ii) for every i, we have $w_i \mathbf{x}_i + b_i \mathbf{1}_i \propto c_1 v_i \mathbf{x}_i + c_2 v_i \mathbf{1}_i$ for $c_1 = \frac{w_i}{v_i}$ and $c_2 = \frac{b_i}{v_i}$. The other five cases are listed in Section 5.7.*

Intuitively, Case (i) refers to when two neurons $i \neq j \in [H]$ have identical activation patterns $\mathbf{1}_i = \mathbf{1}_j$, and so their associated breakpoints have the same orientations $s_i = s_j$ and are in the same data interval $\beta_i, \beta_j \in [x_n, x_{n+1}]$. Since the training loss depends only on $\hat{\mathbf{y}}$, any (infinitesimal) changes in the neurons' parameters (breakpoints or delta-slopes) that leave $\hat{\mathbf{y}}$ (and thus the training loss) unchanged, will yield zero eigenvalues in the Hessian. Case (ii) refers to when $w_i \mathbf{x}_i + b_i \mathbf{1}_i \propto c_1 v_i \mathbf{x}_i + c_2 v_i \mathbf{1}_i$. Recall the scaling transformation $(w_i, b_i, v_i) \mapsto (\alpha_i w_i, \alpha_i b_i, \alpha_i^{-1} v_i)$ that leaves the approximating function $\hat{f}(x)$ (and thus the training loss) invariant, thus generating a 1-dimensional hyperbolic manifold of constant loss. The condition then refers to the *tangent space* of a point $(w_i, b_i, v_i)$ on this curve.

**Lower Bounding the Flatness of the Hessian.** An important question is: How does overparamatrization impact (the spectrum of) the Hessian? The spline parametrization enables us to lower bound the number of zero eigenvalues of the Hessian (e.g. the 'flatness' [15]) as follows.

**Corollary 3.** *Let $\theta^*$ be a critical point of $\ell(\cdot)$ and assume that its data partition is lonely, and at least one neuron is active on all data. Then the fraction of zero eigenvalues of the Hessian at $\theta^*$ is at least $1 - \frac{2N-3}{3H+1} \xrightarrow{H,N \gg 1} 1 - \frac{2}{3\mathcal{O}}$ where $\mathcal{O} \triangleq H/N$ is the overparametrization ratio.*

Intuitively, as overparametrization $H/N$ increases, the number of neurons with shared activation patterns increases, which in turn means many redundant breakpoints between each pair of datapoints, which increases the number of flat directions (zero eigenvalues) of the Hessian.

### 2.4 Gradient Flow Dynamics for Spline Parameters

Gradient Flow (GF) with respect to $\theta_{\text{NN}}$ for our shallow ReLu NN can be written in terms of the spline parameters as a system of ODEs:

$$\dot{\beta}_i = \frac{v_i(t)}{w_i(t)} [\underbrace{\langle \hat{\boldsymbol{\epsilon}}_i(t), \mathbf{1}\rangle}_{\text{net relevant residual}} + \beta_i(t) \underbrace{\langle \hat{\boldsymbol{\epsilon}}_i(t), \mathbf{x}\rangle}_{\text{correlation}}] \tag{5}$$

$$\dot{\mu_i} = w_i^2(t)\left[-\left(1+\left(\frac{v_i(t)}{w_i(t)}\right)^2\right)\langle\hat{\mathbf{e}}_i(t),\mathbf{x}\rangle \right.$$
$$\left. + \beta_i(t)\langle\hat{\mathbf{e}}_i(t),\mathbf{1}\rangle\right] \tag{6}$$

Here $i \in [H]$ indexes all hidden neurons and the initial conditions $\beta_i(0), \mu_i(0) \forall i \in [H]$ must be specified by the initialization used (see Section 5.8 for derivation). Thus, each unit $i \in [H]$ can be interpreted as an independent 'agent' trying to fit an OLS solution to the residuals it can 'see' i.e. agent $i \in [H]$ sees input $n \in [N]$ iff $\hat{a}_{in} = [\![(x_n - \beta_i)_{s_i} > 0]\!]$.

**Impact of Init Scale $\alpha$.** As mentioned previously, the function $\hat{f}$ and the loss is invariant to $\alpha$-scaling transformations of the NN parameters (Section 2.1). However, such scalings do have a large effect on the *initial* NN gradients; specifically, the *initial* learning rate for $\beta_i$ scales as $\frac{1}{\alpha_i^2}$, while that of $\mu_i$ scales approximately as $\alpha_i^2$ (see the Theorem 7). Changing $\alpha$ at initialization has a large impact on the final learned solution. In particular, in the next section we will show how $\alpha$ determines the kind of (implicit) regularization seen in NN training [23].

**Videos of Gradient Descent/Flow Dynamics.** We have developed a BDSO spline parametrization-based visualization. For many of the experiments in this paper, the corresponding videos showing the learning dynamics are available at `http://shorturl.at/tFWZ2`.

### 2.5 A Spline Explanation for Implicit Regularization: Redundant Breakpoints

One of the most useful and perplexing properties of deep neural networks has been that, in contrast to other high capacity function approximators, overparameterizing a neural network does not tend to lead to excessive overfitting [20]. Where does this generalization power come from? What is the mechanism? Much recent work [18, 17] has argued that this so-called implicit regularization (IR) is due to the optimization algorithm itself (i.e. SGD). Using the spline perspective, we confirm this view: IR depends critically on breakpoint and delta-slope learning dynamics. In particular, in the kernel regime $\alpha \to \infty$, breakpoints move very little whereas delta-slopes move a lot, resulting in diffuse populations of breakpoints that distribute curvature. In stark contrast, in the rich regime $\alpha \to 0$ breakpoints move a lot whereas delta-slopes move very little, resulting in tight clusters of breakpoints that concentrate curvature.

**Kernel Regime** We first analyze the so-called kernel regime, inspired by [7] where it is referred to as 'lazy training'. In this section, we omit the overall bias $b_0$.

**Lemma 2.** *Consider the dynamics of gradient flow on $\ell(\cdot)$ started from $\theta_{NN,\alpha} \triangleq (\alpha\mathbf{w}_0, \alpha\mathbf{b}_0, \mathbf{v}_0 = \mathbf{0})$, where $w_i \neq 0 \forall i \in [H]$. In the limit $\alpha \to \infty$, $\boldsymbol{\beta}(t)$ does not change i.e. each breakpoint location is fixed. In this case, the $\theta_{NN}$ model reduces to a (kernel) linear regression:*

$$\hat{\mathbf{y}} = \boldsymbol{\Phi}(\mathbf{x};\boldsymbol{\beta})\boldsymbol{\mu} \tag{7}$$

*where $\boldsymbol{\mu} \in \mathbb{R}^H$ are the regression weights and $\boldsymbol{\Phi}(\mathbf{x};\boldsymbol{\beta}) \in \mathbb{R}^{N\times H}$ are the nonlinear features i.e. $\phi_{ni} \triangleq (x_n - \beta_i)_{s_i}$.*

Using this, we can then adapt a well known result about linear regression (see e.g. [24]):

**Theorem 6.** *Let $\boldsymbol{\mu}^*$ be the converged $\boldsymbol{\mu}$ parameter after gradient flow on the BDSO model Equation* (7) *starting from $\boldsymbol{\mu}_0 = \mathbf{0}$, with $\boldsymbol{\beta}$ held constant. And furthermore suppose that the model perfectly interpolates the training data $\tilde{\ell}(\theta_{BDSO}) = 0$. Then,*

$$\boldsymbol{\mu}^* = \arg\min_{\boldsymbol{\mu}} \|\boldsymbol{\mu}\|_2^2 \text{ s.t. } \mathbf{y} = \boldsymbol{\Phi}(\mathbf{x};\boldsymbol{\beta})\boldsymbol{\mu}.$$

Thus, the case where breakpoint locations are fixed reduces to $\ell_2$-regularized linear regression on the delta-slope parameters $\boldsymbol{\mu}$. Considering $\boldsymbol{\mu}$'s role as the change in slope, i.e. second derivative of $\hat{f}(x;\theta_{\text{BDSO}})$, we are effectively minimizing the sum of the square of the second derivative. If we consider the infinite-width limit, we get the following result:

**Corollary 4.** *Consider the setting of Theorem 6, with the additional assumption that the breakpoints are uniformly spaced, and let $H \to \infty$. Then the learned function $\hat{f}_\infty(x;\boldsymbol{\mu}^*,\boldsymbol{\beta}^*)$ is the global minimizer of*

$$\inf_f \int_{-\infty}^{\infty} f''(x)^2 \, \mathrm{d}x \ \text{ s.t. } y_n = f(x_n) \ \forall n \in [N], \tag{8}$$

*As such,* $\hat{f}_\infty(x; \boldsymbol{\mu}^*, \boldsymbol{\beta}^*)$ *is a natural cubic smoothing spline with* $N$ *degrees of freedom [3].*

This result is the most important overlap with [22], specifically Theorem 5, which is a generalization of Corollary 4 which includes the non-uniform case by replacing Equation (8) with

$$\inf_f \int_{-\infty}^{\infty} \frac{f''(x)^2}{p_0(x)} \, \mathrm{d}x \ \text{ s.t. } y_n = f(x_n) \ \forall n \in [N],$$

where $p_0(\beta)$ is the initial density of breakpoints induced by the specific initialization is used. Thus the initialization has the impact of weighting the curvature of certain locations more than others.

*Explanation for Correlation between Flatness of Minima and Generalization Error.* A key unexplained empirical observation has been that flatter local minima tend to generalize better [15, 21]. Our results above provide an explanation: as overparametrization $\mathcal{O} = H/N$ increases, the flatness of the local minima (as measured by the number of zero eigenvalues) increases (Corollary 3) and the smoothness of the implicitly regularized function (as measured by inverse roughness $\rho(\hat{f}_H) \geq \rho(\hat{f}_\infty)$) also increases.

**Rich/Adaptive Regime** Previous work has shown that the kernel regression regime is insufficient to explain the generalization achieved by modern DL. Instead, the non-convexity of the optimization (e.g. the 'rich' regime, $\alpha \to 0$) must be responsible. The spline parameterization clearly separates the two, with all rich regime effects due to breakpoint dynamics $\beta_i(t)$. Empirically, we discovered an intriguing set of behaviors as $\alpha$ was decreased from $\infty$ towards 0. At intermediate $\alpha$, we found *concentrations* of breakpoints forming near datapoints $x_n$, leading to a final fit that was close to a linear spline interpolation of the data. As $\alpha$ was further reduced, the concentrations of breakpoints become tighter clusters, but only at *some* $x_n$, while breakpoints move past other $x_n$. Clusters at datapoints can remain fixed for some time, before suddenly shifting (all breakpoints in a cluster move together away from $x_n$) or splitting (two groups of $\beta$ form two new sub-clusters that move away from $x_n$ in each direction). These new movements can cause 'smearing', when the cluster loses coherence.

Our theory of the loss surface, its quadratic spline structure with the three types of knots at datapoints (Theorem 3 and Fig. 1), can explain these behaviors. To understand why the rich regime exists, we observe that the gradient of $\ell(\beta_i, \ldots)$ w.r.t. $\beta_i$ scales with $\mu_i$. This means that as long as $\mu_i(t)$ remains small, then loss is less sensitive to variation in $\beta_i$, such that $\beta_i \approx \beta_j$ will have a high probability of sharing the knot types. Low $\alpha$ promotes these conditions, allowing knot types to be stable. Thus, we arrive at the following corollary explaining why we can see concentrations or clusters at some $x_n$, but not all:

**Corollary 5.** *When knot types persist long enough, type (i) knots allow breakpoints to smoothly move past the knot* $x_n$*, type (ii) knots will form repulsors at datapoints, and type (iii) knots form attractors at datapoints.*

See also Lemma 2 of [22], which gives conditions for a data point to be a type (ii) or (iii) knot vs type (i). We believe that the BDSO parameterization makes these results easier to understand and interpret, however.

The knot types are not static; as training progresses, they can change. Type (iii) to (ii) knot transitions result in a breakpoint cluster splitting into two clusters, one on each side of the new repeller knot. Meanwhile Type (iii) to (i) knot transitions result in breakpoints moving past a previously stable cluster. Smearing is caused by these type transitions occurring at slightly different times for different breakpoints. This is confirmed experimentally for lower $\alpha$ (see Supplemental Videos).

Intriguingly, this suggests a possible explanation for why the kernel regime is insufficient: the rich regime enables breakpoint mobility, allowing the NN to adjust the initial breakpoint distribution (and thus, basis of activation patterns). This implies that data-dependent breakpoint initializations may be quite useful (see Table 1 for a simple experiment in this vein).

**Comparison with [22]** Concurrent with and independent of our work, [22] has implicit regularization results in the kernel and rich regimes which parallel our results in this section rather closely. Comparing our results those of Williams *et al.*, the key differences are: (1) our BDSO parametrization has a clear geometric/functional interpretation whereas Williams *et al.*'s canonical parameters are opaque. (2) BDSO generalizes straightforwardly to high dimensions in a conceptually clean manner: oriented breakpoints become oriented breakplanes; it is not clear what the multivariate analogue of the canonical parameters would be. (3) Lemma 2 of [22] is slightly weaker than our Lemma 3 (our conditions delineate attractor/repulsor/passover whereas theirs do not). (4) Our proof techniques are quite different from theirs, with different intuitions and intermediate results, and also simpler. (5) Theorem 5 of [22] is slightly more general form of our Corollary 4, extending to the case of non-uniform breakpoints (though our results can be easily generalized by scaling by breakpoints density). (6) Finally, our results outside of Section 2.5 are not shared. All in all, we feel our results are quite complementary to those of Williams *et al.*

Table 1: Test loss for standard vs uniform breakpoint initialization, on sine and quadratic $\frac{x^2}{2}$

| Init | Sine | Quadratic |
|---|---|---|
| Standard | $4.096 \pm 2.25$ | $.1032 \pm 0404$ |
| Uniform | $2.280 \pm .457$ | $.1118 \pm .0248$ |

# 3 Experiments

**Breakpoint and delta-slope distributions at initialization** We first test our initialization theory against real networks. We initialize fully-connected ReLU networks of varying depths, according to the popular He initializations [12]. Figure 5 shows experimentally measured densities of breakpoints and delta-slopes. Our theory matches the experiments well. The main points to note are that: (i) breakpoints are indeed more highly concentrated around the origin, and that (ii) as depth increases, delta-slopes have lower variance and thus lead to even flatter initial functions. We next ask whether the standard initializations will experience difficulty fitting functions that have significant curvature away from the origin (e.g. learning the energy function of a protein molecule). We train ReLU networks to fit a periodic function ($\sin(x)$), which has high curvature both at and far from the origin. We find that the standard initializations do quite poorly away from the origin, consistent with our theory that breakpoints are essential for modeling curvature. Probing further, we observe empirically that breakpoints cannot migrate very far from their initial location, even if there are plenty of breakpoints overall, leading to highly suboptimal fits. In order to prove that it is indeed the breakpoint density that is causally responsible, we attempt to rescue the poor fitting by using a simple data-dependent initialization that samples breakpoints uniformly over the training data range $[x_{min}, x_{max}]$, achieved by exploiting Equation (1). We train shallow ReLU networks on training data sampled from a sine and a quadratic function, two extremes on the spectrum of curvature. The data shows that uniform breakpoint density rescues bad fits in cases with significant curvature far from the origin, with less effect on other cases, confirming the theory. We note that this could be a potentially useful data-dependent initialization strategy, one that can scale to high dimensions, but we leave this for future work.

**Generalization: Implicit Regularization emerges from Flat Init and Curvature-based Parametrization.** Our theory predicts that IR depends critically upon flatness of the initialization (Theorem 6 and Corollary 4). Here, we test this theory for the case of shallow and deep univariate ReLU nets. We compare training with the standard flat initialization to a 'spiky' initialization, and find that both fit the training data near perfectly, but that the 'spiky' initialization has much worse generalization error (Table 3, Figure 3).

It appears that generalization performance is not automatically guaranteed by GD, but is instead due in part to the flat initializations which are then *preserved* by GD. Our theoretical explanation is simple: integrating the dynamics in Equation (6) yields $\mu_i(t) = \mu_i(0) + \cdots$ and so the initialization's impact remains.

**Impact of Width and Init Variance.** Next, we examine how smoothness (roughness) depends on the width $H$, focusing on settings with large gaps in the training data. We use two discontinuous target functions (shown in Figure 10), leading to a gap in the data, and test how increasing $H$ (with $\alpha$ unchanged) affects the smoothness of $\hat{f}$. We test this on a 'smooth' data gap that is easily fit, as well as a 'sharp' gap, where the fit will require a sharper turn. We trained shallow ReLU networks with varying width $H$ and initial weight variance $\sigma_w$ until convergence, and measured the total roughness of resulting CPWL approximation in the data gaps.

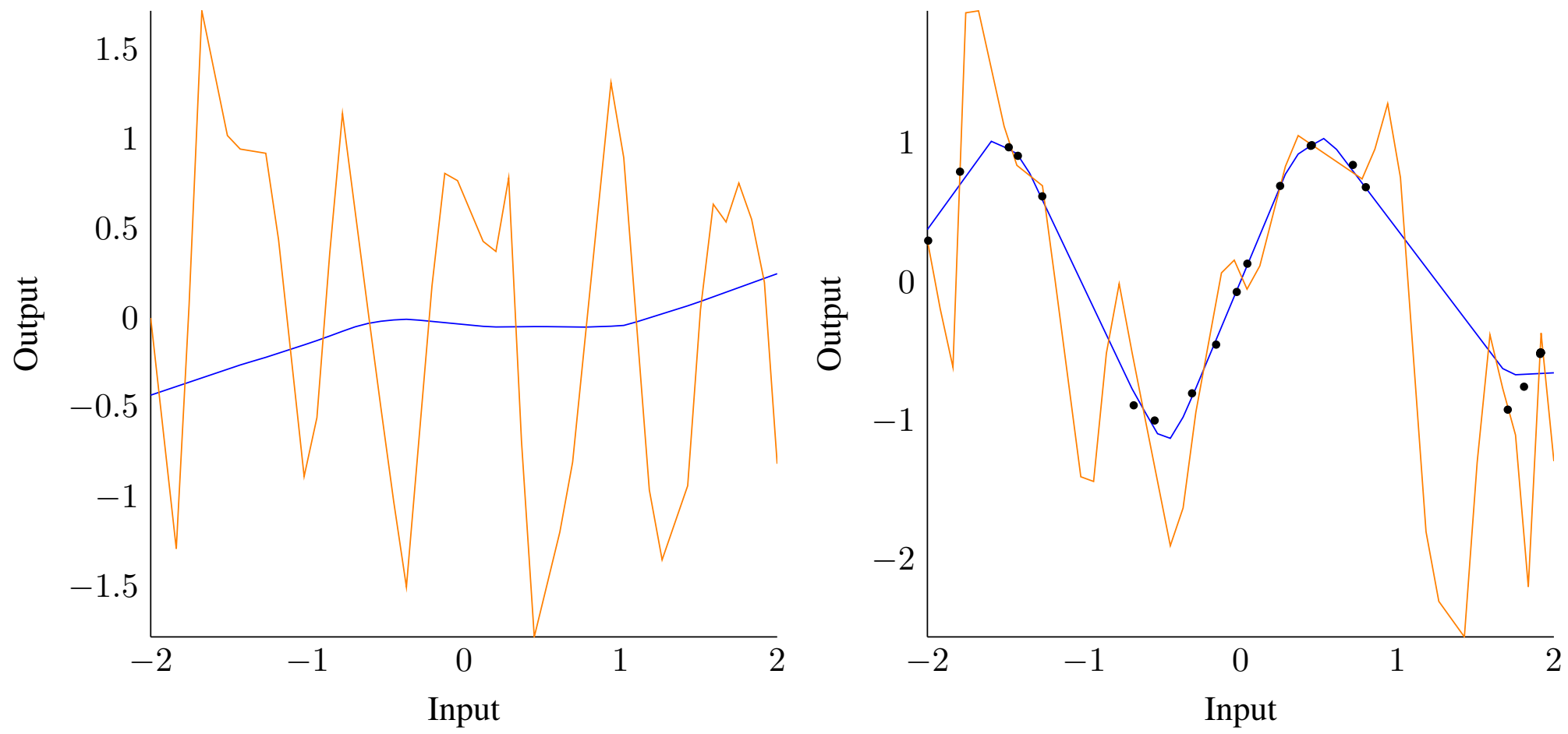

Figure 3: 'Spiky' (orange) and standard initialization (blue), compared before (left) and after (right) training. Note both cases reached similar, very low training set error.

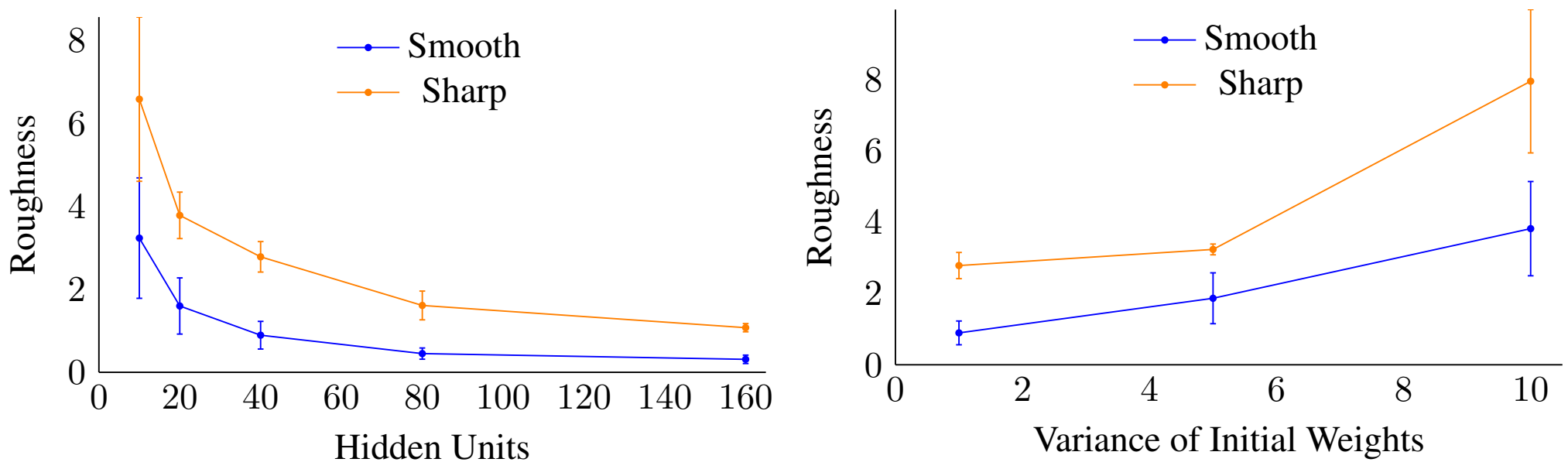

Figure 4: Roughness vs. Width (left) and the variance of the initialization (right) for both data gap cases shown in Figure 10. Each datapoint is the result of averaging over 4 trials trained to convergence.

Figure 4 shows that roughness in the data gaps decreases with width and increases with initial weight variance, confirming our theory. A higher weight variance, and thus rougher initialization, acts in a similar fashion to the 'spiky' initialization, and leads to increased roughness at convergence. In contrast, higher width distributes the curvature 'work' over more units, leading to lower overall roughness.

**Impact of Init Scale $\alpha$** Finally, we explore how changing $\alpha$ impacts IR. Empirically, as $\alpha$ increases from 0 to $\infty$ we see three qualitative regimes: (i) an *under*fitting linear spline, (ii) an interpolating linear spline, and (iii) a roughness-minimizing natural cubic spline. This is quantified in Table 2, where we compare the NN fit to a linear interpolation and a natural cubic spline fit, for varying $\alpha$. We first test in the special case that the initial function approximation is perfectly flat; we find excellent agreement with the linear interpolation and cubic spline fits for $\alpha = 3, 100$ (Uniform initialization) and $\alpha = 10, 100$ (He initialization). The impact of $\alpha$ on IR can be more easily visualized in our Supplemental Videos. In order to gauge the impact of the initialization breakpoint distribution, we also test with a standard He initialization (Cauchy distributed breakpoints, Corollary 1). In this case, we find that generalization error is uniformly higher for all $\alpha$. More strikingly, the $\alpha$ regime (ii) above disappears, as a result of breakpoints being more concentrated around the origin and the initialization roughness being significantly nonzero. This supports the idea that the initial parameter settings, in particular the breakpoint distribution, has a critical impact on the final fit and its IR.

Table 2: Comparison of Neural Network trained to near 0 training loss on random data against linear interpolation and natural cubic splines for varying $\alpha$, with uniform initialization (top) and standard He (bottom). Mean $\pm$ s.d. over 5 random seeds

| $\alpha$ | MAE vs Linear | RMSE vs Linear | MAE vs Cubic | RMSE vs Cubic |
|---|---|---|---|---|
| .1 | .251 ± .077 | .370 ± .11 | .326 ± .12 | .442 ± .16 |
| 1 | .137 ± .060 | .228 ± .074 | .199 ± .084 | .282 ± .12 |
| 3 | .0296 ± 0.0083 | .0749 ± .018 | .117 ± .034 | .158 ± .048 |
| 10 | .122 ± .027 | .157 ± .029 | .0341 ± .012 | .0481 ± .019 |
| 100 | .159 ± .042 | .210 ± .055 | .0299 ± .011 | .0501 ± .024 |
| .1 | .202 ± .079 | .320 ± .13 | .293 ± .11 | .418 ± .16 |
| 1 | .134 ± .064 | .233 ± .11 | .211 ± .10 | .308 ± .15 |
| 3 | .132 ± .065 | .239 ± .12 | .209 ± .089 | .329 ± .15 |
| 10 | .115 ± .046 | .163 ± .061 | .0884 ± .052 | .149 ± .11 |
| 100 | .161 ± .048 | .212 ± .055 | .0556 ± .015 | .0828 ± .021 |

Taken together, our experiments strongly support that a smooth, flat initialization and overparametrization are both responsible for the phenomenon and strength of IR, while the initialization weight scale $\alpha$ critically determines the type of IR.

**Conclusions.** We show that examining neural networks in spline space enabled us to glean new theoretical and practical insights. The spline view highlights the importance of initialization: a smooth initial approximation is required for a smooth final solution. Fortunately, existing inits used in DL practice approximate this property. In spline space, we also achieve a surprisingly simple and transparent view of the loss surface, its critical points, its Hessian, and the phenomenon of overparameterization. It clarifies how increasing width relative to data size leads w.h.p. to lonely data partitions, which in turn are more likely to reach global minima. The spline view also allows us to elucidate the phenomenon of implicit regularization, and how it arises due to overparametrization and the init scale $\alpha$.

Looking forward, there are still many questions to answer from the spline perspective: How does depth affect the expressive power, learnability, and IR? What kinds of regularization are induced in the rich regime and how do modern deep nets take advantage of them? How can data-dependent inits of the breakpoints help rescue/improve the performance of GD? Can we design new global learning algorithms inspired based on breakpoint (re)allocation? We believe the BDSO perspective can help answer these questions.

Additionally, we believe that the BDSO parameterization can be extended to multivariate inputs: for $D$-dimensional inputs, you can write $\hat{f}(\mathbf{x};\theta_{\text{NN}})$ as

$$\hat{f}(\mathbf{x};\theta_{\text{NN}}) = \sum_{i=1}^{H} v_i \left(\langle \mathbf{w}_i, \mathbf{x}\rangle + b_i\right)_+ + b_0,$$

where the input weights are represented as $D$-dimensional vectors $\mathbf{w}_i$. Using the reparameterization $\eta_i \triangleq v_i\|\mathbf{w}_i\|_2$, $\boldsymbol{\xi}_i \triangleq \frac{\mathbf{w}_i}{\|\mathbf{w}_i\|_2}$, $\gamma_i \triangleq \frac{-b_i}{\|\mathbf{w}_i\|_2}$, the representation becomes

$$\hat{f}(\mathbf{x};\theta_{\text{BDSO}}) = \sum_{i=1}^{H} \eta_i \left(\langle \boldsymbol{\xi}, \mathbf{x}\rangle - \gamma_i\right)_+ + b_0,$$

where $\eta_i$ is a "delta-slope"[1] parameter, and $(\boldsymbol{\xi}_i, \gamma_i)$ parameterize the orientation and signed distance from the origin of a $D-1$-dimensional "oriented breakplane" (generalizing the 0-dimensional left-or-right oriented breakpoint represented by $(s_i, \beta_i)$ in the 1-dimensional case). Generalizing our results to this parameterization is ongoing work.

[1]For $D = 1$, we have $\eta_i = v_i|w_i|$, differing from the delta-slope $\mu_i = v_i w_i$ used in this paper by the sign of $w_i$

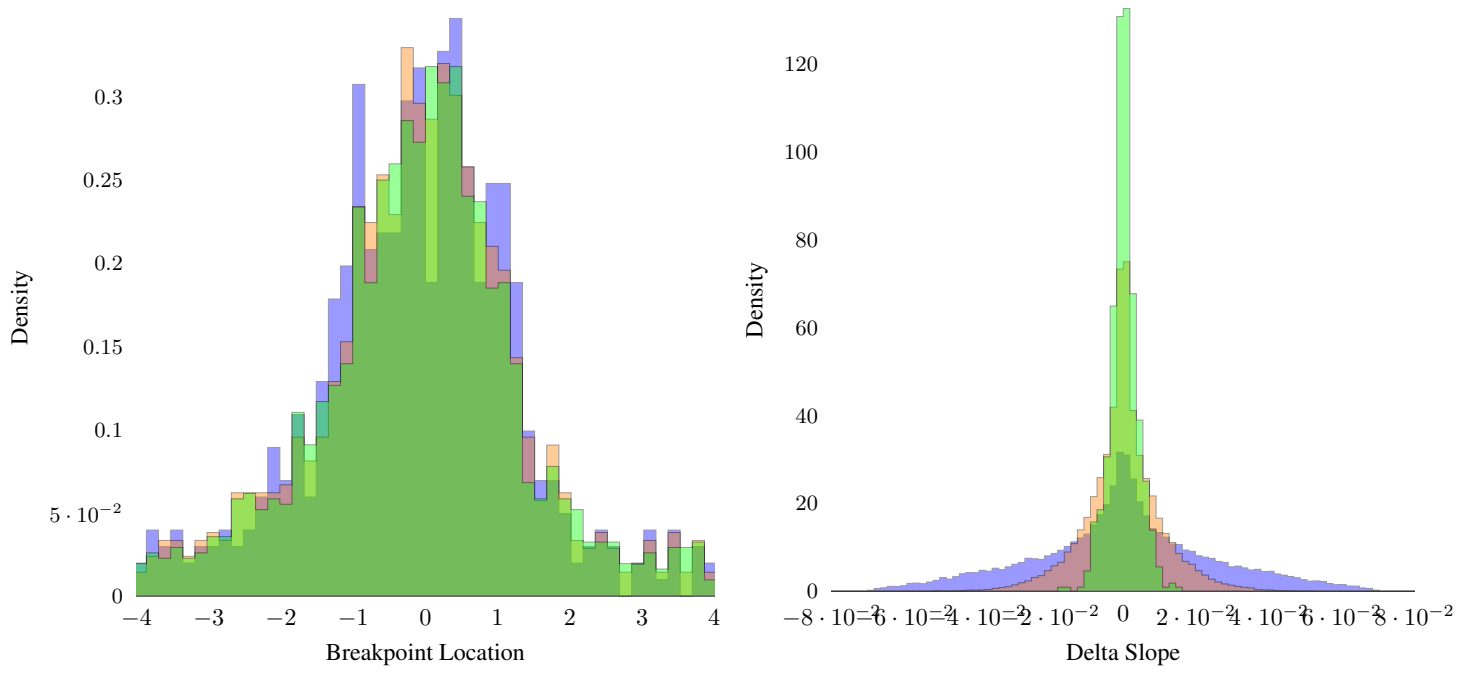

Figure 5: Left: Breakpoint distribution for a He initialization across a 3 layer network. Right: Delta-slope distribution for a He initialization across a 3 layer network.

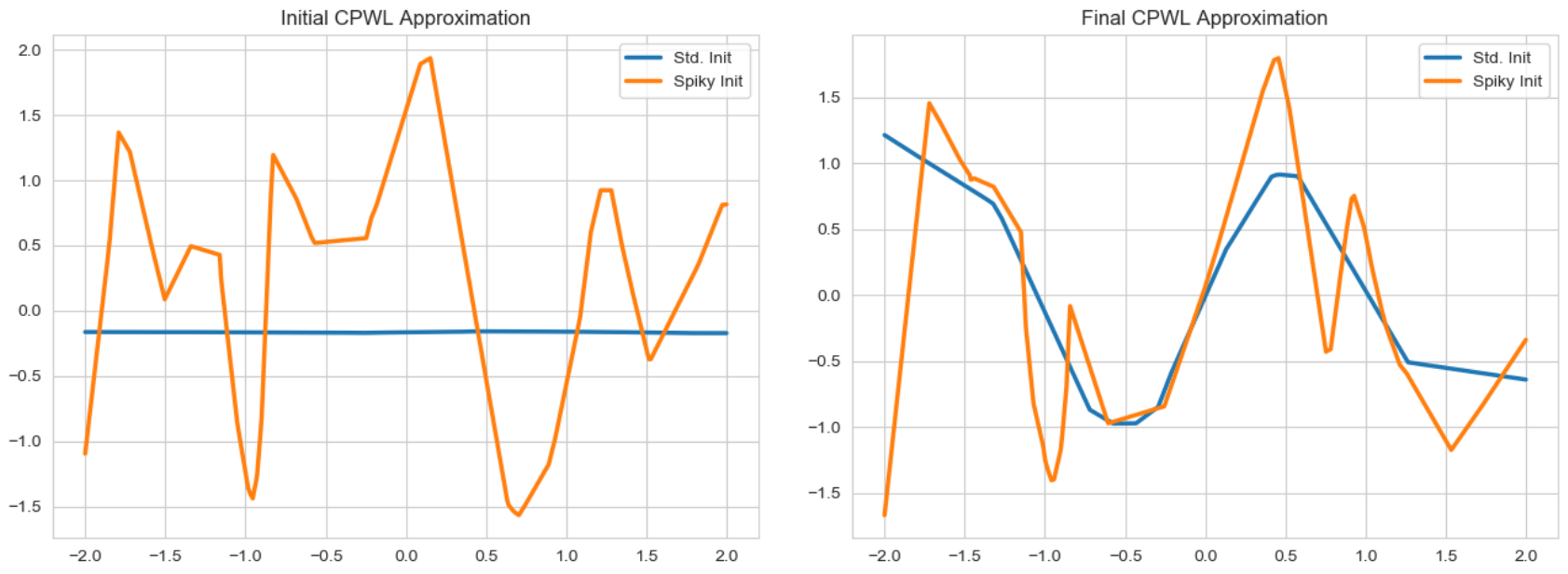

Figure 6: 'Spiky' (orange) and standard initialization (blue), compared before training (left) and post-training (right) using a deep network

# 4 Experimental Details

## 4.1 Uniform Initialization

Trained on a shallow, 21 unit FC ReLU network. Trained on function over the interval [-2,2]. Learning rate = 5e-5, trained via GD over 10000 epochs. Compared against pytorch default of He initialization. Training data sampled uniformly every .01 of the target interval. Each experiment was run 5 times, with results reported as mean $\pm$ standard deviation. Breakpoints y values were taken from the original standard initialization for the uniform initialization plus a small random noise term N(0,.01), making initial condition within the target interval nearly identical.

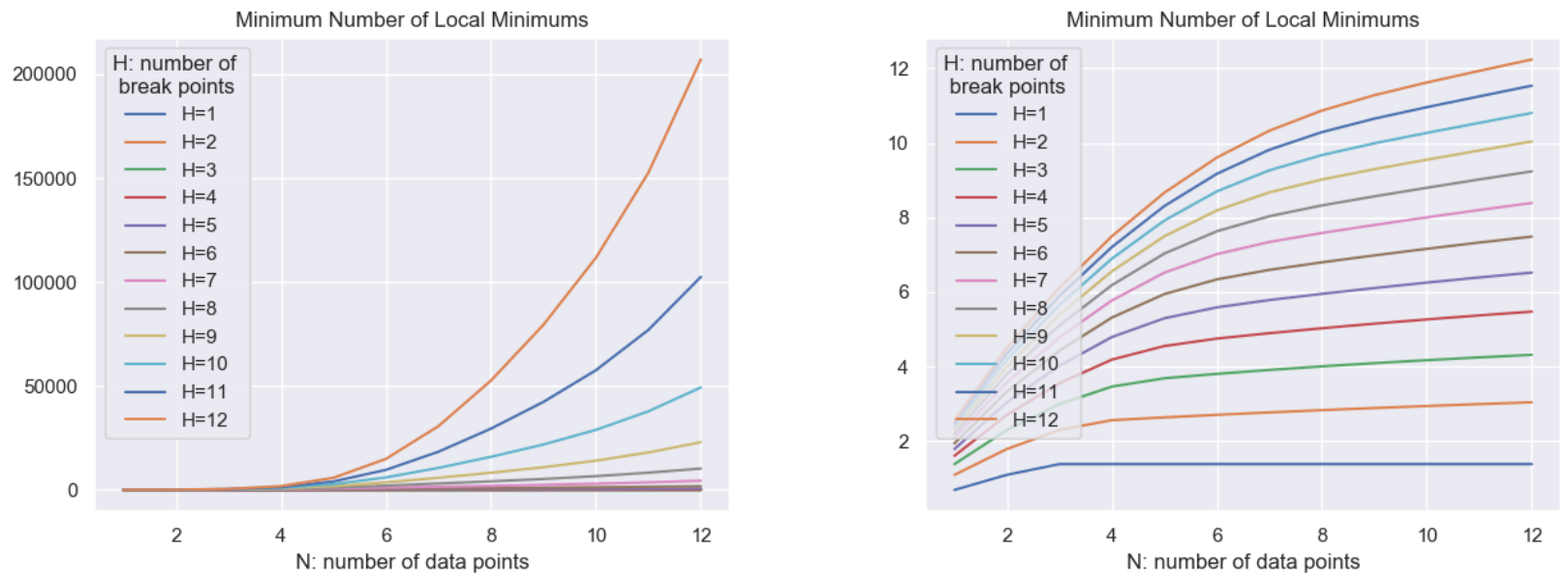

Figure 7: Growth in the (minimum) amount of local minima, as a function of the number of breakpoints and datapoints. Right plot is identical, but with log scaling

Table 3: Comparison of testing loss of various network shallow and deep networks with a standard vs 'spiky' initialization

| Function | Shallow | Spiky Shallow | Deep | Spiky Deep |
|---|---|---|---|---|
| Sine | $42.95 \pm 6.406$ | $157.5 \pm 60.27$ | $31.48 \pm 7.078$ | $122.0 \pm 128.2$ |
| Arctan | $0.013 \pm 0.077$ | $2.499 \pm 1.257$ | $0.980 \pm 0.936$ | $32.570 \pm 26.100$ |
| Sawtooth | $156.9 \pm 12.45$ | $150.1 \pm 61.48$ | $148.1 \pm 8.755$ | $198.0 \pm 170.9$ |
| Cubic | $3.608 \pm 1.683$ | $136.7 \pm 124.1$ | $56.77 \pm 98.91$ | $191.6 \pm 114.1$ |
| Quadratic | $3.559 \pm 4.553$ | $150.6 \pm 49.00$ | $1.741 \pm 1.296$ | $46.02 \pm 19.42$ |
| Exp | $.6509 \pm .5928$ | $181.1 \pm 75.36$ | $1.339 \pm 1.292$ | $54.50 \pm 37.77$ |

### 4.2 Spiky Initialization

For a shallow ReLU network, we can test a 'spiky' initialization by exactly solving for network parameters to generate a given arbitrary CPWL function. This network initialization is then compared against a standard initialization, and trained against a smooth function with a small number of training datapoints. Note that in a 1D input space we need a small number of training datapoints to create a situation similar to that of the sparsity caused by high dimensional input, and to allow for testing generalization between datapoints.

For a deep ReLU network, it is more difficult to exactly solve for a 'spiky' initialization. Instead, we train a network to approximate an arbitrary CPWL function, and call those trained network parameters the 'spiky' initialization. Once again, the 'spiky' initialization has near identical training performance, hitting all datapoints, but has noticeably worse generalization performance. **Detailed Experimental**

**Procedure** Shallow version trained on a 21 unit FC ReLU Network. Deep version trained on a deep, 5-layer network with 4 hidden layers of width 8. In both cases, the 'spiky' initialization was a 20 - breakpoint CPWL function, with $y_n \sim$ Uniform$([-2, 2])$. In the deep case, the spiky model was initialized with the same weights as the non-spiky model, and then pre-trained for 10,000 epochs to fit the CPWL. After that, gradient descent training proceeded on both models for 20,000 epochs, with all training having learning rate 1e-4. Training data was 20 random points in the range [-2,2], while the testing data (used to measure generalization) was spaced uniformly at every $\Delta x = .01$ of the target interval of the target function.

In the shallow case, there was no pre-training, as the 'spiky' model was directly set to be equal to the CPWL. In the shallow model, training occurred for 20,000 epochs. All experiment were run over 5 trials, and values in table are reported as mean $\pm$ standard deviation. Base shallow learning rate was 1e-4 using gradient descent method, with learning rate divided by 5 for the spiky case due to the initialization method generating larger weights. Despite differing learning rates, both models had similar training loss curves and similar final training loss values, e.g. for sine, final training loss was .94 for spiky and 1.02 for standard. Functions used were $\sin(x), \arctan(x)$, a sawtooth function from [-2,2] with minimum value of -1 at the endpoints, and 4 peaks of maximum value 1, cubic $\frac{x^3}{4} + \frac{x^2}{2} - \frac{x}{2}$, quadratic $\frac{x^2}{2}$, and $\exp(.5x)$ Note GD was chosen due to the strong theoretical focus of this paper - similar results were obtained using ADAM optimizer, in which case no differing learning rates were necessary.

### 4.3 Explaining the Need for Overparameterization

One possible explanation for why overparameterization is helpful is that it makes lonely partitions more likely to occur at initialization and remain lonely at subsequent times, resulting in a lonely final partition which can be trivially fit with 0 training loss. Figure 11 confirms the hypothesis above, for both our custom uniform and standard He initializations. If gradient descent training on neural networks must be overparameterized to succeed, it implies that GD is highly inefficient with its parameters, leading us to ask the question: under what conditions will GD be successful? Empirically, it has been observed that neural nets must be massively overparameterized (relative to the number of parameters needed to express the underlying function) in order to ensure good training performance. We focus on the case of a univariate fully connected shallow ReLU network. A univariate input (i) enables us to use our theory, (ii) allows for visualization of the entire learning trajectory, and (iii) enables direct comparison with existing globally (near-)optimal algorithms for fitting PWL functions. The latter include the Dynamic Programming algorithm (DP, [5]), and a very fast greedy approximation known as Greedy Merge (GM, [1]). How do these algorithms compare to GD, across different target function classes, in terms of training loss relative to the number of pieces/hidden units? Taking the spline-based view allows us to directly compare neural network performance to these PWL approximation algorithms. For a quadratic function (e.g. with high curvature, requiring many pieces), we find that the globally optimal DP algorithm can quickly reduce training error to near 0 with order 100 pieces. The GM algorithm, a relaxation of the DP algorithm, requires slightly higher pieces, but requires significantly less computational power. On the other hand all variants of GD (vanilla, Adam, SGD w/ BatchNorm) all require far more pieces to reduce error below a target threshold, and may not even monotonically decrease error with number of pieces. Interestingly, we observe a strict ordering of optimization quality with Adam outperforming BatchNorm SGD outperforming Vanilla GD. These results (Figure 8) show how inefficient GD is with respect to parameters, requiring orders of magnitude more for similar performance to exact or approximate PWL fitting algorithms.

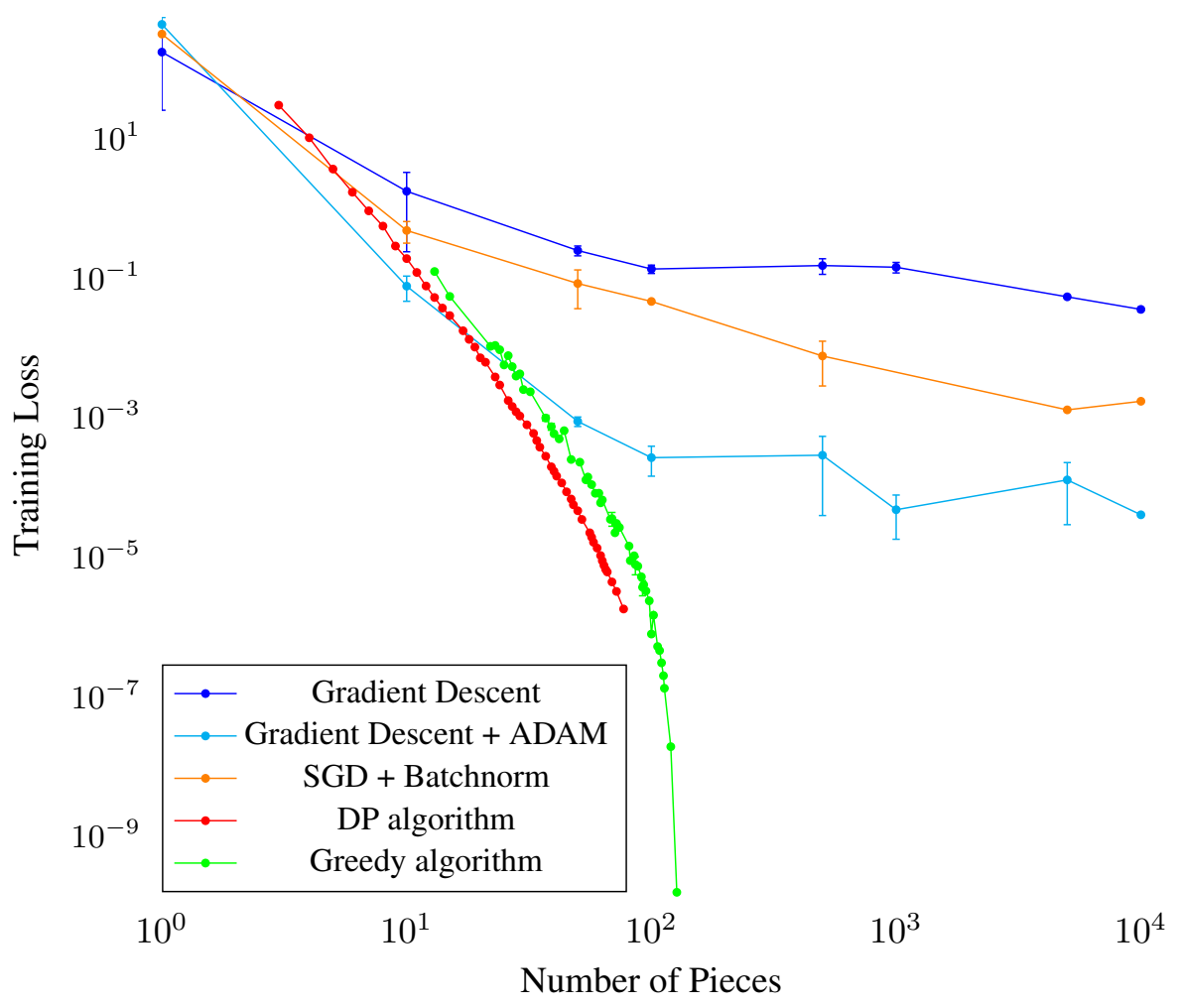

Figure 8: Training loss vs number of pieces ($\propto$ number of parameters) for various algorithms fitting a CPWL function to a quadratic.

Finally, we compare a standard He initialization vs our uniform initialization on a very small task. The He initialization leads to lonely partitions, preventing it from achieving training loss near 0 for $\alpha = 100$, while the uniform initialization with all lonely partitions is able to do so. These results can be seen in our Supplemental Videos.

### 4.4 Comparing Neural Network to Linear or Cubic Splines

In this section, we used a non-standard initialization in order to better compare with our theory. In particular, we set $w$ to randomly be plus or minus 1 with equal probability, $b$ to be a uniform partition of our x-range multiplied by the sign of $w$ s.t. the breakpoints uniformly tile our x-range. $v$ was multiplied by 0, leading to a perfectly flat initialization - note that as this is a shallow network, gradient can still flow through this after the first gradient descent update. Networks had hidden size $H = 1000$, and $N = 21$ datapoints were chosen with $x$ values uniformly spaced from $[-2, 2]$, with an additional $x$ datapoint at 3 and -3. $y$ values for all datapoints were randomly chosen between $[-2, 2]$ independently. Prior to training, we constructed a linear interpolation of the data, as well as a natural cubic smoothing spline using the R method `smooth.spline`. $\alpha$ modification took place once before training began. Training was done using Adam, with a learning rate of 3e-5 for $\alpha \leq 1$, otherwise learning rate was divided by 5 to ensure that the networks had converged after 50,000 training epochs. The final neural network fit was compared against the spline fits over a fine grid with a $\Delta x$ of .01 from -2 to 2. We considered the extra datapoints at $\pm$ 3 to act as 'pins' providing boundary conditions - without these, the neural network fits would still interpolate all interior datapoints, but potentially with different slopes through the first and last datapoints, leading to a sub-optimal fit.

### 4.5 Supplemental Videos

We generate supplemental videos, using the settings in the above section at a fixed seed. We show both the general lower learning rate, and a faster learning rate version when necessary to show certain dynamics (i.e. for lower $\alpha$). These videos are available at `http://shorturl.at/tFWZ2`.

### 4.6 Varying $\alpha$

We look at a specific zoomed in portion of the trials above on a fixed seed, training to convergence, and show the results in Figure 9. This gives an intuitive picture of what varying $\alpha$ does, with too low giving an underfit, moderate giving a linear interpolation, and higher approaching the natural cubic spline.

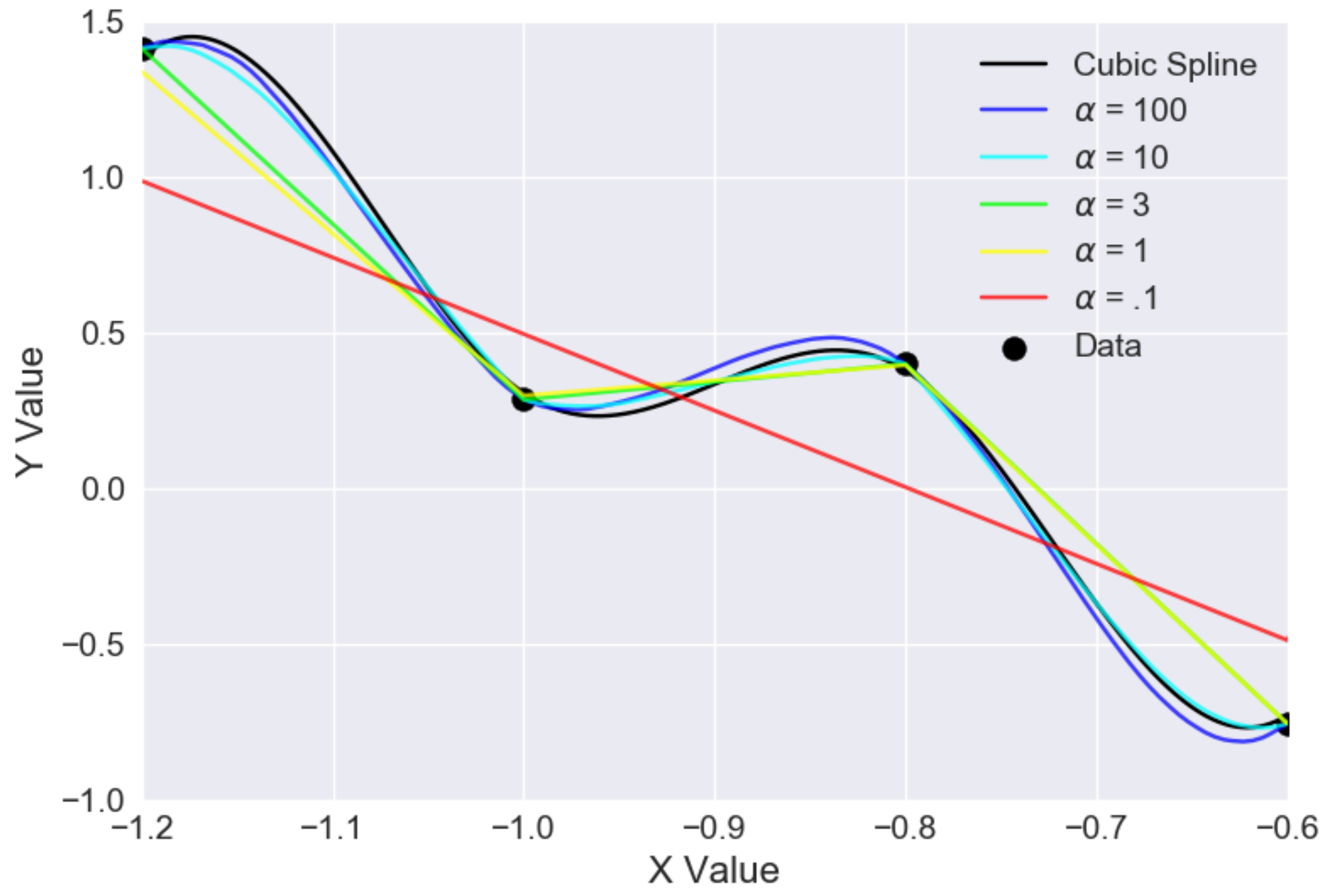

Figure 9: Effect of $\alpha$, using same experimental protocol as $Table$ 2, zoomed in on a small area of a single seed.

## 5 Proofs of Theoretical Results

### 5.1 Reparametrization from ReLu Network to Piecewise Linear Function

*Proof of Equation* (1)*:*

$$\begin{aligned}
\hat{f}_{\theta,H}(x) &= \sum_{i=1}^{H} v_i \phi(w_i x + b_i) \\
&= \sum_{i=1}^{H} v_i (w_i x + b_i) [\![ w_i x + b_i > 0 ]\!] \\
&= \sum_{i=1}^{H} v_i w_i (x - \beta_i) \begin{cases} [\![ x > \beta_i ]\!] & w_i > 0 \\ [\![ x < \beta_i ]\!] & w_i < 0 \end{cases} \quad \text{where } \beta_i \triangleq -\frac{b_i}{w_i}
\end{aligned}$$

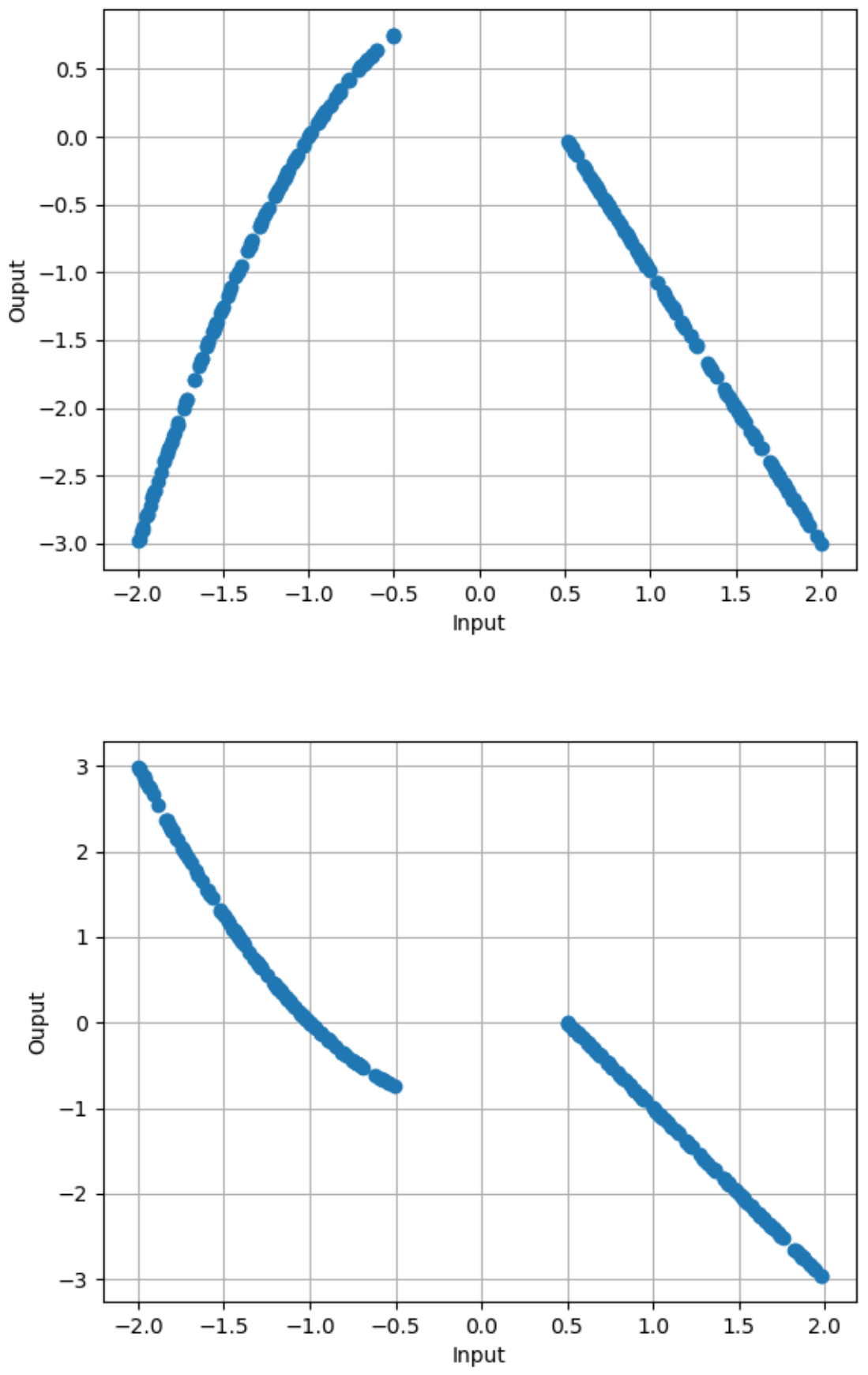

Figure 10: Training data sampled from two ground truth functions, one smoothly (left) and the other sharply (right) discontinuous, each with a data gap at $[-0.5, 0.5]$.

$$= \sum_{i=1}^{H} \mu_i (x - \beta_i) \begin{cases} [\![x > \beta_i]\!] & w_i > 0 \\ [\![x < \beta_i]\!] & w_i < 0 \end{cases} \quad \text{where } \mu_i \triangleq v_i w_i$$

This gives us Equation (1), as desired. Let the subscripts $p$, $q$ denote the parameters sorted by $\beta_p$ value. In this setting, let $\beta_0 \triangleq -\infty$, and $\beta_{H+1} \triangleq \infty$. Then,

$$= \sum_{p=0}^{H} \left( \sum_{\substack{q=1 \\ w_q > 0}}^{p} \mu_q (x - \beta_q) + \sum_{\substack{q=p+1 \\ w_q < 0}}^{H} \mu_q (x - \beta_q) \right) [\![\beta_p \leq x < \beta_p]\!]$$

$$= \sum_{p=0}^{H} \left( \sum_{\substack{q=1 \\ w_q > 0}}^{p} (\mu_q x - \mu_q \beta_q) + \sum_{\substack{q=p+1 \\ w_q < 0}}^{H} (\mu_q x - \mu_q \beta_q) \right) [\![\beta_p \leq x < \beta_{p+1}]\!]$$

$$= \sum_{p=0}^{H} \left( \sum_{\substack{q=1 \\ w_q > 0}}^{p} \mu_q x - \sum_{\substack{q=1 \\ w_q > 0}}^{p} \mu_q \beta_q + \sum_{\substack{q=p+1 \\ w_q < 0}}^{H} \mu_q x - \sum_{\substack{q=p+1 \\ w_q < 0}}^{H} \mu_q \beta_q \right) [\![\beta_p \leq x < \beta_{p+1}]\!]$$

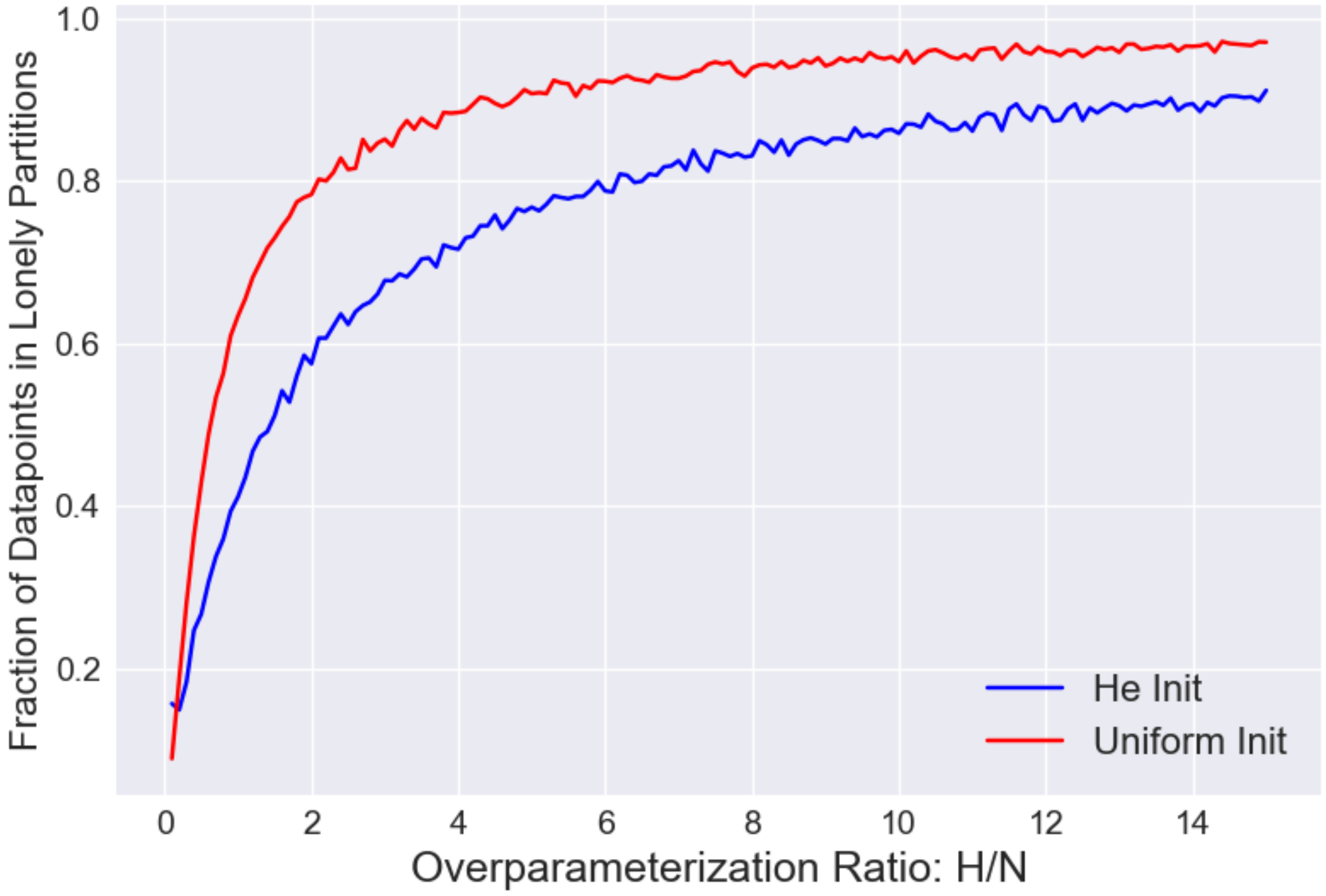

Figure 11: Percentage of datapoints which are in a lonely partition as a function of overparameterization ratio $\frac{H}{N}$ for both a standard (He) and uniform init. Massive overparameterization leads w.h.p. to lonely partitions.

$$= \sum_{p=0}^{H} \left( \left( \sum_{\substack{q=1 \\ w_q>0}}^{p} \mu_q \right) x - \sum_{\substack{q=1 \\ w_q>0}}^{p} \mu_q \beta_q + \left( \sum_{\substack{q=p+1 \\ w_q<0}}^{H} \mu_q \right) x - \sum_{\substack{q=p+1 \\ w_q<0}}^{H} \mu_q \beta_q \right) [\![\beta_p \le x < \beta_{p+1}]\!]$$

$$= \sum_{p=0}^{H} \left( \underbrace{\left( \sum_{\substack{q=1 \\ w_q>0}}^{p} \mu_q + \sum_{\substack{q=p+1 \\ w_q<0}}^{H} \mu_q \right)}_{\triangleq m_p} x - \underbrace{\left( \sum_{\substack{q=1 \\ w_q>0}}^{p} \mu_q \beta_q + \sum_{\substack{q=p+1 \\ w_q<0}}^{H} \mu_q \beta_q \right)}_{\triangleq \gamma_p} \right) [\![\beta_p \le x < \beta_{p+1}]\!]$$

$$= \sum_{p=0}^{H} (m_p x - \gamma_p) [\![\beta_p \le x < \beta_{p+1}]\!]$$

This gives us Equation (1), as desired. □

### 5.2 Random Initialization in Function Space

**Lemma 3.** *Suppose* $(b_i, w_i, v_i)$ *are initialized independently with densities* $f_B(b_i)$*,* $f_W(w_i)$*, and* $f_V(v_i)$*. Then, the density of* $(\beta_i, \mu_i)$ *is given by*

$$f_{\beta,\mu}(\beta_i, \mu_i) = \int_{-\infty}^{\infty} f_B(\beta_i u) f_W(u) f_V(\frac{\mu_i}{u}) \, \mathrm{d}u \, .$$

*Proof.* Suppose $(b_i, w_i, v_i)$ are initialized i.i.d. from a distribution with density $f_{B,W,V}(b_i, w_i, v_i)$. Then, we can derive the density of $(\beta_i, \mu_i)$ by considering the invertable continuous transformation given by $(\beta_i, \mu_i, u) = g(b_i, w_i, v_i) = (b_i/w_i, v_i|w_i|, w_i)$, where $g^{-1}(\beta_i, \mu_i, u) = (\beta_i u, u, \mu_i/|u|)$. The density of $(\beta_i, \mu_i, u)$ is given by $f_{B,M,V}(\beta_i u, u, \mu_i/|u|)|J|$, where $J$ is the Jacobian determinant

of $g^{-1}$. Then, we have $J = -\operatorname{sgn} w_i$ and $|J| = 1$. The density of $(\beta_i, \mu_i)$ is then derived by integrating out the dummy variable $u$: $f_{\beta,\mu}(\beta_i, \mu_i) = \int_{-\infty}^{\infty} f_{B,W,V}(\beta_i u, u, \frac{\mu_i}{u})\,\mathrm{d}u$. If $(b_i, w_i, v_i)$ are independent, this expands to $\int_{-\infty}^{\infty} f_B(\beta_i u) f_W(u) f_V(\frac{\mu_i}{u})\,\mathrm{d}u$. $\square$

### 5.3 Gaussian Initialization in Function

**Theorem 1(a).** *Consider a fully connected ReLU neural net with scalar input and output, and a single hidden layer of width $H$. Let the weights and biases be initialized randomly according to a zero-mean Gaussian or Uniform distribution. Then, under an independent Gaussian initialization,*

$$p_{\beta,\mu}(\beta_i, \mu_i) = \frac{1}{2\pi\sigma_v\sqrt{\sigma_b^2 + \sigma_w^2\beta_i^2}} \exp\left[-\frac{|\mu_i|\sqrt{\sigma_b^2 + \sigma_w^2\beta_i^2}}{\sigma_b\sigma_v\sigma_w}\right]$$

*Proof.* Starting with Lemma 3,

$$\begin{aligned}
f_{\beta,\mu}(\beta, \mu) &= \int_{-\infty}^{\infty} f_B(\beta_i u) f_W(u) f_V(\frac{\mu_i}{u})\,\mathrm{d}u \\
&= \int_{-\infty}^{\infty} \frac{1}{\sqrt{2\pi\sigma_b^2}} e^{-\frac{(\beta u)^2}{2\sigma_b^2}} \frac{1}{\sqrt{2\pi\sigma_w^2}} e^{-\frac{u^2}{2\sigma_w^2}} \frac{1}{\sqrt{2\pi\sigma_v^2}} e^{-\frac{(\mu/u)^2}{2\sigma_v^2}} du \\
\text{(Sympy)} &= \begin{cases} \dfrac{\exp\left[-\frac{\mu\sqrt{\sigma_b^2+\sigma_w^2(\beta)^2}}{\sigma_b\sigma_v\sigma_w}\right]}{2\pi\sigma_v\sqrt{\sigma_b^2+\sigma_w^2(\beta)^2}} & \mu > 0 \\ \text{unknown} & \text{otherwise} \end{cases}
\end{aligned}$$

but the integrand is even in $\mu$, giving

$$= \frac{\exp\left[-\frac{|\mu|\sqrt{\sigma_b^2+\sigma_w^2(\beta)^2}}{\sigma_b\sigma_v\sigma_w}\right]}{2\pi\sigma_v\sqrt{\sigma_b^2 + \sigma_w^2(\beta)^2}}$$

$\square$

**Corollary 1(a).** *Consider the same setting as Theorem 1. In the case of an independent Gaussian initialization,*

$$\begin{aligned}
p_\beta(\beta_i) &= \mathit{Cauchy}\left(\beta_i; 0, \frac{\sigma_b}{\sigma_w}\right) = \frac{\sigma_b\sigma_w}{\pi\left(\sigma_w^2\beta_i^2 + \sigma_b^2\right)} \\
p_\mu(\mu_i) &= \frac{1}{2\pi\sigma_v\sigma_w} G_{0,2}^{2,0}\left(\frac{\mu_i^2}{4\sigma_v^2\sigma_w^2}\middle| 0, 0\right) = \frac{1}{\pi\sigma_v\sigma_w} K_0\left(\frac{|\mu_i|}{\sigma_v\sigma_w}\right) \\
p_{\mu|\beta}(\mu_i|\beta_i) &= \mathit{Laplace}\left(\mu_i; 0, \frac{\sigma_b\sigma_v\sigma_w}{\sqrt{\sigma_b^2 + \sigma_w^2\beta_i^2}}\right) = \frac{\sqrt{\sigma_b^2 + \sigma_w^2\beta_i^2}}{2\sigma_b\sigma_v\sigma_w} \exp\left[-\frac{|\mu_i|\sqrt{\sigma_b^2 + \sigma_w^2\beta_i^2}}{\sigma_b\sigma_v\sigma_w}\right],
\end{aligned}$$

*where $G_{pq}^{nm}(\cdot|\cdot)$ is the Meijer G-function and $K_\nu(\cdot)$ is the modified Bessel function of the second kind.*

*Proof.* Marginalizing out $\mu$ from the joint density in Sympy returns the desired $f_\beta(\beta)$ from above. Sympy cannot compute the other marginal, so we verify it by hand:

$$\begin{aligned}
f_\mu(\mu) = \int_{-\infty}^{\infty} \frac{\exp\left[-\frac{|\mu|\sqrt{\sigma_b^2+\sigma_w^2\beta^2}}{\sigma_b\sigma_v\sigma_w}\right]}{2\pi\sigma_v\sqrt{\sigma_b^2 + \sigma_w^2\beta^2}} dx &= \frac{1}{2\pi\sigma_v} \int_{-\infty}^{\infty} \frac{\exp\left[-\frac{|\mu|\sqrt{\sigma_b^2+\sigma_w^2\beta^2}}{\sigma_b\sigma_v\sigma_w}\right]}{\sqrt{\sigma_b^2 + \sigma_w^2\beta^2}} dx \\
\left(\phi(\beta) = \frac{\beta}{\sigma_w}\right) \quad &= \frac{1}{2\pi\sigma_v \underbrace{\sigma_w}_{\phi'(\beta)}} \int_{-\infty}^{\infty} \frac{\exp\left[\frac{-|\mu|\sqrt{\sigma_b^2+\beta^2}}{\sigma_b\sigma_v\sigma_w}\right]}{\sqrt{\sigma_b^2 + \beta^2}} dx
\end{aligned}$$

from [10], Eq. 3.462.20, we have

$$K_0(ab) = \int_0^\infty \frac{\exp\left(-a\sqrt{\beta^2+b^2}\right)}{\sqrt{\beta^2+b^2}}\,\mathrm{d}\beta \qquad [\operatorname{Re} a > 0, \operatorname{Re} b > 0]$$

$$\text{(integrand is even in } \beta\text{)} = \frac{1}{2}\int_{-\infty}^\infty \frac{\exp\left(-a\sqrt{\beta^2+b^2}\right)}{\sqrt{\beta^2+b^2}}\,\mathrm{d}\beta \qquad [\operatorname{Re} a > 0, \operatorname{Re} b > 0]$$

applying this with $a = \frac{|\mu|}{\sigma_b\sigma_v\sigma_w}$ and $b = \sigma_b$,

$$\frac{1}{2\pi\sigma_v\sigma_w}\int_{-\infty}^\infty \frac{\exp\left[\frac{-|\mu|\sqrt{\sigma_b^2+\beta^2}}{\sigma_b\sigma_v\sigma_w}\right]}{\sqrt{\sigma_b^2+\beta^2}}\,\mathrm{d}\beta = \frac{1}{\pi\sigma_v\sigma_w}K_0\left(\frac{|\mu|}{\sigma_v\sigma_w}\right)$$

as desired. We can then use these densities to derive the conditional:

$$\begin{aligned} f_\mu(\mu|\beta) &= \frac{\sqrt{\sigma_b^2+\sigma_w^2(\beta)^2}\exp\left[-\frac{|\mu|\sqrt{\sigma_b^2+\sigma_w^2(\beta)^2}}{\sigma_b\sigma_v\sigma_w}\right]}{2\sigma_b\sigma_v\sigma_w} \\ &= \text{Laplace}\left(\mu; 0, \frac{\sigma_b\sigma_v\sigma_w}{\sqrt{\sigma_b^2+\sigma_w^2(\beta)^2}}\right). \end{aligned}$$

□

### 5.4 Uniform Initialization in Function Space

**Theorem 1(b).** *Consider a fully connected ReLU neural net with scalar input and output, and a single hidden layer of width $H$. Let the weights and biases be initialized randomly according to a zero-mean Gaussian or Uniform distribution. Then, under an independent Uniform initialization,*

$$p_{\beta,\mu}(\beta_i,\mu_i) = \frac{[\![|\mu_i| \le \min\{\frac{a_b a_v}{|\beta_i|}, a_w, a_v\}]\!]}{4a_b a_w a_v}\left(\min\{\frac{a_b}{|\beta_i|}, a_w\} - \frac{|\mu_i|}{a_v}\right)$$

*Proof.* Starting with Lemma 3,

$$\begin{aligned} f_{\beta,\mu}(\beta,\mu) &= \int_{-a_w}^{a_w} f_B(\beta u) f_W(u) f_V(\mu/u)\,\mathrm{d}u \\ &= \int_{-a_w}^{a_w} \frac{1}{2a_b}[\![-a_b \le \beta u \le a_b]\!]\frac{1}{2a_w}[\![-a_w \le u \le a_w]\!]\frac{1}{2a_v}[\![-a_v \le \mu/u \le a_v]\!]\,\mathrm{d}u \\ &= \int_{-a_w}^{a_w} \frac{1}{2a_b}[\![-a_b/|\beta| \le u \le a_b/|\beta|]\!]\frac{1}{2a_w}[\![-a_w \le u \le a_w]\!]\frac{1}{2a_v}[\![u \le -|\mu|/a_v \vee u \ge |\mu|/a_v]\!]\,\mathrm{d}u \\ &= \int_{-a_w}^{a_w} \frac{1}{8a_b a_w a_v}[\![-\min\{a_b/|\beta|, a_w\} \le u \le -|\mu|/a_v \vee |\mu|/a_v \le u \le \min\{a_b/|\beta|, a_w\}]\!] \\ &\qquad \times [\![|\mu| \le a_b a_v/|\beta|]\!]\,\mathrm{d}u \\ &= \frac{[\![|\mu| \le a_b a_v/|\beta|]\!]}{8a_b a_w a_v}\int_{-a_w}^{a_w} [\![-\min\{a_b/|\beta|, a_w\} \le u \le -|\mu|/a_v \vee |\mu|/a_v \le u \le \min\{a_b/|\beta|, a_w\}]\!]\,\mathrm{d}u \\ &= \frac{[\![|\mu| \le a_b a_v/|\beta|]\!]}{4a_b a_w a_v}\int_0^{a_w} [\![|\mu|/a_v \le u \le \min\{a_b/|\beta|, a_w\}]\!]\,\mathrm{d}u \\ &= \frac{[\![|\mu| \le a_b a_v/|\beta|]\!]}{4a_b a_w a_v}\left(\min\{a_b/|\beta|, a_w\} - |\mu|/a_v\right)[\![-a_w a_v \le \mu \le a_w a_v]\!] \end{aligned}$$

as desired. □

**Corollary 1(b).** *Consider the same setting as Theorem 1. In the case of an independent Uniform initialization,*

$$p_\beta(\beta_i) = \frac{1}{4a_b a_w}\left(\min\left\{\frac{a_b}{|\beta_i|}, a_w\right\}\right)^2$$

$$p_\mu(\mu_i) = \frac{[\![-a_w a_v \le \mu_i \le a_w a_v]\!]}{2a_w a_v} \log \frac{a_w a_v}{|\mu_i|}$$

$$p_{\mu|\beta}(\mu_i|\beta_i) = \mathit{Tri}(\mu_i; a_v \min\{a_b/|\beta_i|, a_w\}) = \frac{[\![|\mu_i| \le a_v \min\{a_b/|\beta_i|, a_w\}]\!]}{a_v \min\{a_b/|\beta_i|, a_w\}} \left(1 - \frac{|\mu_i|}{a_v \min\{a_b/|\beta_i|, a_w\}}\right),$$

*where $\mathit{Tri}(\cdot; a)$ is the symmetric triangular distribution with base $[-a, a]$ and mode $0$.*

*Proof.* Beginning with the marginal of $\beta_i$,

$$\begin{aligned}
f_\beta(\beta) &= \int f_{\beta,\mu}(\beta,\mu)\,\mathrm{d}\mu \\
&= \int_{-\infty}^{\infty} \frac{[\![|\mu| \le a_b a_v/|\beta|]\!]}{4a_b a_w a_v} \left(\min\{a_b/|\beta|, a_w\} - |\mu|/a_v\right) [\![-a_w a_v \le \mu \le a_w a_v]\!]\,\mathrm{d}\mu \\
&= \int_{-a_w a_v}^{a_w a_v} \frac{[\![|\mu| \le a_b a_v/|\beta|]\!]}{4a_b a_w a_v} \left(\min\{a_b/|\beta|, a_w\} - |\mu|/a_v\right)\mathrm{d}\mu \\
&= \frac{1}{2a_b a_w a_v} \int_0^{a_w a_v} [\![\mu \le a_b a_v/|\beta|]\!] \left(\min\{a_b/|\beta|, a_w\} - \mu/a_v\right)\mathrm{d}\mu \\
&= \frac{1}{2a_b a_w a_v} \left(\int_0^{a_w a_v} [\![\mu \le a_b a_v/|\beta|]\!] \min\{a_b/|\beta|, a_w\}\,\mathrm{d}\mu - \int_0^{a_w a_v} [\![\mu \le a_b a_v/|\beta|]\!] \mu/a_v\,\mathrm{d}\mu\right) \\
&= \frac{1}{2a_b a_w a_v} \left(\min\{a_b/|\beta|, a_w\} \int_0^{a_w a_v} [\![\mu \le a_b a_v/|\beta|]\!]\,\mathrm{d}\mu - \frac{1}{a_v} \int_0^{a_w a_v} [\![\mu \le a_b a_v/|\beta|]\!] \mu\,\mathrm{d}\mu\right) \\
&= \frac{1}{2a_b a_w a_v} \left(\min\{a_b/|\beta|, a_w\} \min\{a_w a_v, a_b a_v/|\beta|\} - \frac{1}{a_v} \int_0^{\min\{a_w a_v, a_b a_v/|\beta|\}} \mu\,\mathrm{d}\mu\right) \\
&= \frac{1}{2a_b a_w a_v} \left(\min\{a_b/|\beta|, a_w\} \min\{a_w a_v, a_b a_v/|\beta|\} - \frac{1}{2a_v} \left(\min\{a_w a_v, a_b a_v/|\beta|\}\right)^2\right) \\
&= \frac{1}{2a_b a_w a_v} \left(a_v \left(\min\{a_b/|\beta|, a_w\}\right)^2 - \frac{1}{2a_v} \left(a_v \min\{a_w, a_b/|\beta|\}\right)^2\right) \\
&= \frac{1}{2a_b a_w a_v} \left(a_v \left(\min\{a_b/|\beta|, a_w\}\right)^2 - \frac{a_v}{2} \left(\min\{a_w, a_b/|\beta|\}\right)^2\right) \\
&= \frac{1}{4a_b a_w} \left(\min\{a_b/|\beta|, a_w\}\right)^2
\end{aligned}$$

as desired. Then,

$$\begin{aligned}
f_\mu(\mu) &= \int f_{\beta,\mu}(\beta,\mu)\,\mathrm{d}\beta \\
&= \int_{-\infty}^{\infty} \frac{[\![|\mu| \le a_b a_v/|\beta|]\!]}{4a_b a_w a_v} \left(\min\{a_b/|\beta|, a_w\} - |\mu|/a_v\right) [\![-a_w a_v \le \mu \le a_w a_v]\!]\,\mathrm{d}\beta \\
&= \frac{[\![-a_w a_v \le \mu \le a_w a_v]\!]}{4a_b a_w a_v} \int_{-\infty}^{\infty} [\![|\mu| \le a_b a_v/|\beta|]\!] \left(\min\{a_b/|\beta|, a_w\} - |\mu|/a_v\right)\mathrm{d}\beta \\
&= \frac{[\![-a_w a_v \le \mu \le a_w a_v]\!]}{4a_b a_w a_v} 2 \int_0^{\infty} [\![|\mu| \le a_b a_v/\beta]\!] \left(\min\{a_b/\beta, a_w\} - |\mu|/a_v\right)\mathrm{d}\beta \\
&= \frac{[\![-a_w a_v \le \mu \le a_w a_v]\!]}{4a_b a_w a_v} 2 \int_0^{\infty} [\![\beta \le a_b a_v/|\mu|]\!] \left(\min\{a_b/\beta, a_w\} - |\mu|/a_v\right)\mathrm{d}\beta \\
&= \frac{[\![-a_w a_v \le \mu \le a_w a_v]\!]}{4a_b a_w a_v} 2 \int_0^{a_b a_v/|\mu|} \min\{a_b/\beta, a_w\} - |\mu|/a_v\,\mathrm{d}\beta \\
&= \frac{[\![-a_w a_v \le \mu \le a_w a_v]\!]}{4a_b a_w a_v} 2 \left(\int_0^{\min\{a_b a_v/|\mu|, a_b/a_w\}} a_w - |\mu|/a_v\,\mathrm{d}\beta\right. \\
&\qquad\qquad \left. + \int_{\min\{a_b a_v/|\mu|, a_b/a_w\}}^{a_b a_v/|\mu|} a_b/\beta - |\mu|/a_v\,\mathrm{d}\beta\right)
\end{aligned}$$

$$= \frac{[\![-a_w a_v \le \mu \le a_w a_v]\!]}{4a_b a_w a_v} 2\left(\int_0^{a_b/a_w} a_w - |\mu|/a_v \, \mathrm{d}\beta + \int_{a_b/a_w}^{a_b a_v/|\mu|} a_b/\beta - |\mu|/a_v \, \mathrm{d}\beta\right)$$
$$= \frac{[\![-a_w a_v \le \mu \le a_w a_v]\!]}{4a_b a_w a_v} 2\left((a_b/a_w)(a_w - |\mu|/a_v) + \int_{a_b/a_w}^{a_b a_v/|\mu|} a_b/\beta \, \mathrm{d}\beta - \int_{a_b/a_w}^{a_b a_v/|\mu|} |\mu|/a_v \, \mathrm{d}\beta\right)$$
$$= \frac{[\![-a_w a_v \le \mu \le a_w a_v]\!]}{4a_b a_w a_v} 2\Bigg[(a_b/a_w)(a_w - |\mu|/a_v) + \int_{a_b/a_w}^{a_b a_v/|\mu|} a_b/\beta \, \mathrm{d}\beta - (|\mu|/a_v)(a_b a_v/|\mu| - a_b/a_w)\Bigg]$$
$$= \frac{[\![-a_w a_v \le \mu \le a_w a_v]\!]}{4a_b a_w a_v} 2\left((a_b/a_w)(a_w - |\mu|/a_v) + a_b \log \frac{a_b a_v/|\mu|}{a_b/a_w} - (|\mu|/a_v)(a_b a_v/|\mu| - a_b/a_w)\right)$$
$$= \frac{[\![-a_w a_v \le \mu \le a_w a_v]\!]}{4a_b a_w a_v} 2a_b \log \frac{a_b a_v/|\mu|}{a_b/a_w}$$
$$= \frac{[\![-a_w a_v \le \mu \le a_w a_v]\!]}{2a_w a_v} \log \frac{a_b a_v/|\mu|}{a_b/a_w}$$
$$= \frac{[\![-a_w a_v \le \mu \le a_w a_v]\!]}{2a_w a_v} \log \frac{a_w a_v}{|\mu|}$$

as desired. We can then use these densities to derive the conditional:

$$f_\mu(\mu_i|\beta_i) = \frac{\frac{[\![|\mu_i| \le a_b a_v/|\beta_i|]\!]}{4a_b a_w a_v} (\min\{a_b/|\beta_i|, a_w\} - |\mu_i|/a_v) [\![-a_w a_v \le \mu_i \le a_w a_v]\!]}{\frac{1}{4a_b a_w} (\min\{a_b/|\beta_i|, a_w\})^2}$$
$$= \frac{\frac{[\![|\mu_i| \le a_b a_v/|\beta_i|]\!][\![-a_w a_v \le \mu_i \le a_w a_v]\!]}{a_v} (\min\{a_b/|\beta_i *|, a_w\} - |\mu_i|/a_v)}{(\min\{a_b/|\beta_i|, a_w\})^2}$$
$$= \frac{[\![|\mu_i| \le a_v \min\{a_b/|\beta_i|, a_w\}]\!]}{a_v \min\{a_b/|\beta_i|, a_w\}} \left(1 - \frac{|\mu_i|}{a_v \min\{a_b/|\beta_i|, a_w\}}\right).$$

□

**Remarks.** Note that the marginal distribution on $\mu_i$ is the distribution of a product of two independent random variables, and the marginal distribution on $\beta_i$ is the distribution of the ratio of two random variables. For the Gaussian case, the marginal distribution on $\mu_i$ is a symmetric distribution with variance $\sigma_v^2 \sigma_w^2$ and excess Kurtosis of 6. For the Uniform case, the marginal distribution of $\beta_i$ is a symmetric distribution with no finite higher moments. The marginal distribution of $\mu_i$ is a symmetric distribution with bounded support and variance $\frac{2a_w^3 a_v^3}{9}$ and excess Kurtosis of $\frac{81}{50a_w a_v} - 3$. The conditional distribution of $\mu_i$ given $\beta_i$ is a symmetric distribution with bounded support and variance $\frac{(a_v \min\{a_b/|\beta_i|, a_w\})^2}{6}$ and excess Kurtosis of $-\frac{3}{5}$.

### 5.5 Roughness of Random Initialization

**Theorem 2.** *Consider the initial roughness $\rho_0$ under a Gaussian initialization. In the He initialization, we have that the tail probability is given by*

$$\mathbb{P}[\rho_0 - \mathbb{E}[\rho_0] \ge \lambda] \le \frac{1}{1 + \frac{\lambda^2 H}{128}},$$

*where $\mathbb{E}[\rho_0] = 4$. In the Glorot initialization, we have that the tail probability is given by*

$$\mathbb{P}[\rho_0 - \mathbb{E}[\rho_0] \ge \lambda] \le \frac{1}{1 + \frac{\lambda^2 (H+1)^4}{128H}},$$

*where $\mathbb{E}[\rho_0] = \frac{4H}{(H+1)^2} = O\left(\frac{1}{H}\right)$.*

*Proof.* Using the moments of the delta-slope distribution computed in Theorem 1, Corollary 1, and above, we can compute:

$$\mathbb{E}[\rho_0] = \sum_{i=1}^{H} \mathbb{E}[\mu_i^2] = \sum_{i=1}^{H} \mathrm{Var}[\mu_i] + \mathbb{E}[\mu_i]^2 = H\sigma_v^2 \sigma_w^2$$

$$\begin{aligned}\mathrm{Var}[\rho_0] &= \sum_{i=1}^{H} \mathrm{Var}[\mu_i^2] = H\,\mathrm{Var}[\mu_i^2] = H(\mathbb{E}[\mu_i^4] - \mathbb{E}[\mu_i^2]^2) \\ &= H(9(\sigma_v\sigma_w)^4 - \sigma_v^4\sigma_w^4) = 8H\sigma_v^4\sigma_w^4\end{aligned}$$

Applying the two initializations, we have

$$\begin{aligned}\mathbb{E}[\rho_0^{\mathrm{He}}] &= H\frac{2}{H}\frac{2}{1} = 4 \\ \mathrm{Var}[\rho_0^{\mathrm{He}}] &= 8H\frac{4}{H^2}\frac{4}{1} = \frac{128}{H} \\ \mathbb{E}[\rho_0^{\mathrm{Glorot}}] &= H\frac{2}{H+1}\frac{2}{1+H} = \frac{4H}{(H+1)^2} = O\left(\frac{1}{H}\right) \\ \mathrm{Var}[\rho_0^{\mathrm{Glorot}}] &= 8H\frac{4}{(H+1)^2}\frac{4}{(1+H)^2} = \frac{128H}{(H+1)^4} = O\left(\frac{1}{H^3}\right)\end{aligned}$$

By applying Cantelli's theorem, we get the stated tail probabilities. □

### 5.6 Loss Surface in the Spline Parametrization

**Theorem 3.** *The loss function $\tilde{\ell}(\theta_{BDSO} = (\boldsymbol{\beta}, \boldsymbol{\mu}), \mathbf{s})$ is a continuous piecewise quadratic spline. Furthermore, the 1-dimensional slice $\tilde{\ell}(\beta_i; \boldsymbol{\beta}_{-i}, \boldsymbol{\mu}, \mathbf{s})$ is also a continuous piecewise quadratic spline in $\beta_i$ with knots at datapoints $\{x_n\}_{n=1}^N$. Let $p_1(\beta_i)$ ($p_2(\beta_i)$) be the left (right) piece at knot $x_n$, which both have positive curvature, and let $m_j \triangleq \arg\min p_j(\beta_i)$. Then, with measure 1, the knots $x_n$ fall into one of four types as shown in Figure 1: (Type I) $m_1, m_2 < x_n$, or $x_n < m_1, m_2$, (Type II) $m_1 < x_n < m_2$ (Type III) $m_2 < x_n < m_1$.*

*Proof.* First, consider the following function:

$$g(\boldsymbol{\beta}, \boldsymbol{\mu}) = \sum_{n=1}^{N}\left(\sum_{i=1}^{H} \mu_i\,(x_n - \beta_i) - y_n\right)^2.$$

As a sum of squares of linear terms, this is clearly quadratic in $(\boldsymbol{\beta}, \boldsymbol{\mu})$. Then, note that the loss

$$\tilde{\ell}(\boldsymbol{\beta}, \boldsymbol{\mu}, \mathbf{s}) = \sum_{n=1}^{N}\left(\sum_{i=1}^{H} \mu_i\,(x_n - \beta_i)_{s_i} - y_n\right)^2$$

differs only in the addition of the $(\cdot)_{s_i}$ operation, which converts each linear term to a continuous piecewise linear term. Thus, $\tilde{\ell}(\cdot)$ is continuous piecewise quadratic in $(\boldsymbol{\beta}, \boldsymbol{\mu})$. Furthermore, holding $\boldsymbol{\mu}$ and $\mathbf{s}$ constant, we still have the composition of a quadratic and piecewise linear function. In particular, the piecewise linear terms have their knots at $\beta_i = x_n$, which are inherited by $\tilde{\ell}(\cdot)$.

Finally, because $\tilde{\ell}(\cdot)$ is of the form $\sum_{n=1}^{N} \epsilon_n^2$, it has nonnegative curvature everywhere, so $m_1$ and $m_2$ are well-defined for each knot. Then, we can analyze the loss surface under the various orderings of $m_1, m_2$, and $x_n$. We may ignore the case that $m_j = x_n$ as this is an event of measure 0. It is straightforward to collect the remaining cases into the three types depicted in Figure 1. □

**Theorem 4.** *Consider some arbitrary $\theta_{NN}^*$ such that (i) every datapoint is on the active side of* some *neuron, and (ii) there exists at least one neuron that is active on all data. Then, $\theta_{NN}^*$ is a critical point of $\tilde{\ell}(\cdot)$ if and only if for all pieces $p \in [P]$, $\hat{f}(\cdot; \theta_{NN}^*)|_{\pi_p} = m_p x + \gamma_p$ is an Ordinary Least Squares (OLS) fit of the training data contained in piece p.*

*Proof.* Let $\theta_{\mathrm{NN}}^*$ be as hypothesized. Then, $\theta_{\mathrm{NN}}^*$ is a critical point of $\tilde{\ell}(\cdot)$ iff $\frac{\partial \tilde{\ell}(\cdot)}{\partial \theta_{\mathrm{NN}}}\Big|_{\theta_{\mathrm{NN}}^*} = \mathbf{0}$. Expanding, we have that the right hand sides of Equation (4) are all 0. Without loss of generality, assume there is no $i \in [H]$ such that $v_i = w_i = b_i = 0$. Simplifying under our assumptions, we get

$$\langle \hat{\boldsymbol{\epsilon}}, \mathbf{1} \rangle = 0$$

$$\langle \hat{\boldsymbol{\epsilon}}_i, \mathbf{1} \rangle = 0 \quad \forall i \in [H]$$
$$\langle \hat{\boldsymbol{\epsilon}}_i, \mathbf{x} \rangle = 0 \quad \forall i \in [H]$$

The latter two lines can be written as

$$\Big( \underset{N\times H}{\overline{\boldsymbol{\epsilon}}} \odot \underset{N\times H}{\hat{\mathbf{A}}} \Big)^\top \underset{N\times 2}{\tilde{\mathbf{X}}} = \underset{H\times 2}{\mathbf{0}},$$

where $\overline{\boldsymbol{\epsilon}}$ is $H$ copies of the $N$-dimensional vector $\hat{\boldsymbol{\epsilon}}$, $\hat{a}_{ni} = [\![ w_i x_n + b_i > 0 ]\!]$ is the masking matrix that selects the relevant $\epsilon_n$ for each neuron $i$, and $\tilde{\mathbf{X}} = [\mathbf{1} \quad \mathbf{X}]$. Because the columns of $\overline{\boldsymbol{\epsilon}}$ are identical, we can swap the dot and Hadamard products:

$$= \hat{\mathbf{A}}^\top \Big( \underset{N\times 2}{\overline{\boldsymbol{\epsilon}}'} \odot \tilde{\mathbf{X}} \Big)$$
$$\triangleq \hat{\mathbf{A}}^\top \hat{\mathbf{r}},$$

where $\overline{\boldsymbol{\epsilon}}'$ contains just 2 copies of $\hat{\boldsymbol{\epsilon}}$.

Thus, $\theta^*_{\text{NN}}$ is a critical point iff $\hat{\mathbf{r}}$ is in the null space of $\hat{\mathbf{A}}^\top$. We therefore proceed by analyzing $\hat{\mathbf{A}}^\top$. The $i^{\text{th}}$ row of $\hat{\mathbf{A}}^\top$ is given by $\mathbf{1}_i$; let us assume that the rows are sorted by the corresponding $\beta_i$, and the columns are sorted by the corresponding $x_n$ value. Then, each row will be either consist of a block of 0s and a block of 1s. To illustrate this, consider an example with eight datapoints and five breakpoints:

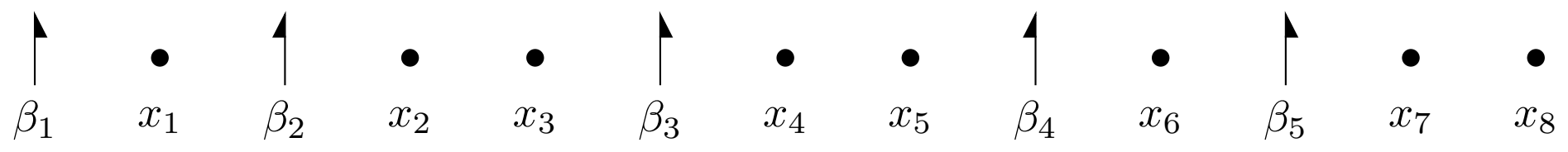

where filled circles denotes a datapoint, and a "flag" denotes a breakpoint, where the direction of the "flag" indicates the facing $s_i$ (e.g. $s_1 = +1$ so that neuron 1 is right-facing and therefore active on all data). Then, we have

$$\hat{\mathbf{A}}_{\text{ex}}^\top = \begin{bmatrix} 1 & 1 & 1 & 1 & 1 & 1 & 1 & 1 \\ 1 & 0 & 0 & 0 & 0 & 0 & 0 & 0 \\ 0 & 0 & 0 & 1 & 1 & 1 & 1 & 1 \\ 1 & 1 & 1 & 1 & 1 & 0 & 0 & 0 \\ 0 & 0 & 0 & 0 & 0 & 0 & 1 & 1 \end{bmatrix}.$$

Using the all-1s vector corresponding to the neuron active on all data, we can then apply elementary row operations to convert rows with leading 1s to have leading 0s. In our example, this gives

$$\hat{\mathbf{A}}_{\text{ex}}^\top = \begin{bmatrix} 1 & 1 & 1 & 1 & 1 & 1 & 1 & 1 \\ 0 & 1 & 1 & 1 & 1 & 1 & 1 & 1 \\ 0 & 0 & 0 & 1 & 1 & 1 & 1 & 1 \\ 0 & 0 & 0 & 0 & 0 & 1 & 1 & 1 \\ 0 & 0 & 0 & 0 & 0 & 0 & 1 & 1 \end{bmatrix}.$$

Because the rows are sorted by breakpoint, and each row's transition from 0s to 1s happens at the index of the first datapoint greater than that row's corresponding breakpoint, there are exactly two cases: (i) row $i+1$ is equal to row $i$ (when the two breakpoints land between the same pair of datapoints), or (ii) row $i+1$ will have some additional 0s corresponding to the datapoints between the two breakpoints. If we iterate from the last row to the first and apply row-wise differences (i.e., subtracting row $i+1$ from row $i$), we get for case (i) a row of all 0s, and for case (ii) a row which is 1 for each datapoint between the two breakpoints:

$$\hat{\mathbf{A}}_{\text{ex}}^\top = \begin{bmatrix} 1 & 0 & 0 & 0 & 0 & 0 & 0 & 0 \\ 0 & 1 & 1 & 0 & 0 & 0 & 0 & 0 \\ 0 & 0 & 0 & 1 & 1 & 0 & 0 & 0 \\ 0 & 0 & 0 & 0 & 0 & 1 & 0 & 0 \\ 0 & 0 & 0 & 0 & 0 & 0 & 1 & 1 \end{bmatrix}.$$

In this form, each row of $\hat{\mathbf{A}}^\top$ is given by the piece indicator $\mathbf{1}_p$, given by $\mathbf{1}_{p,n} = [\![x_n \in [\beta_p, \beta_{p+1})]\!]$. Returning to our goal of proving $\hat{\mathbf{A}}^\top \hat{\mathbf{r}} = \mathbf{0}$, we see that this now reduces to

$$\langle \hat{\boldsymbol{\epsilon}}_p, \mathbf{1} \rangle = 0 \quad \forall p \in [H]$$
$$\langle \hat{\boldsymbol{\epsilon}}_p, \mathbf{x} \rangle = 0 \quad \forall p \in [H],$$

which is exactly the condition that $\hat{f}(\cdot; \theta_{\text{NN}})|_{\pi_p} = m_p x + \gamma_p$ is an OLS fit of the data in piece $p$, for $p \in [H]$. Note that there are $P = H + 1$ pieces, but by assuming that there is at least one breakpoint active on the data (and thus outside the data), the "missing" piece is guaranteed to have no data in it, and thus vacuously be an OLS fit. Finally, the condition $\langle \hat{\boldsymbol{\epsilon}}, \mathbf{1} \rangle = 0$ can be expressed as $\sum_{p=1}^{P} \langle \hat{\boldsymbol{\epsilon}}_p, \mathbf{1} \rangle = \sum_{p=1}^{P} 0 = 0$.

□

**Lemma 1.** *For any lonely partition $\Pi$, there are infinitely many parameter settings $\theta_{BDSO}$ that induce $\Pi$ and are global minima with $\tilde{\ell}(\theta_{BDSO}) = 0$. Furthermore, in the overparametrized regime $H \geq cN$ for some constant $c \geq 1$, the total number of lonely partitions, and thus a lower bound on the total number of* global *minima of $\tilde{\ell}$ is $\binom{H+1}{N} = O(N^{cN})$.*

*Proof.* Note that each linear piece $p$ has two degrees of freedom (slope and intercept). By way of induction, start at (say) the left-most piece. If there is a datapoint in this piece, choose an arbitrary slope and intercept that goes through it; otherwise, choose an arbitrary slope and intercept. At each subsequent piece, we can use one degree of freedom to ensure continuity with the previous piece, and use one degree of freedom to match the data (if there is any). Counting the number of lonely partitions reduces to counting the number of possible ways of allocating $N$ datapoints (balls) into $H + 1$ pieces (urns) with at most one datapoint per piece. □

### 5.7 The Gradient and Hessian of the Loss $\ell(\theta_{\text{NN}})$

**Theorem 5.** *Let $\theta^*$ be a critical point of $\ell(\cdot)$ such that $\beta_i^* \neq x_n$ for all $i \in [H]$ and for all $n \in [N]$. Then the Hessian $\mathbf{H}_\ell(\theta^*)$ is the positive semi-definite Gram matrix of the set of $3H + 1$ vectors*

$$\mathcal{B} \triangleq \{v_i \mathbf{x}_i, w_i \mathbf{x}_i + b_i \mathbf{1}_i, v_i \mathbf{1}_i\}_{i=1}^{H} \cup \{\mathbf{1}\},$$

*as shown in Equation* (3)*. Thus, $\mathbf{H}_\ell(\theta^*)$ is positive definite iff the vectors of this set are linearly independent.*

*Proof.* We begin by deriving the first and second derivatives of $\hat{\mathbf{f}}$:

$$\begin{aligned}
\frac{\partial \hat{\mathbf{f}}}{\partial b_0} &= \frac{\partial}{\partial b_0} \left( \sum_{i=1}^{H} v_i \phi(w_i \mathbf{x} + b_i \mathbf{1}) + b_0 \mathbf{1} \right) \\
&= \mathbf{1} \\
\frac{\partial \hat{\mathbf{f}}}{\partial w_i} &= \frac{\partial}{\partial w_i} \left( \sum_{i=1}^{H} v_i \phi(w_i \mathbf{x} + b_i \mathbf{1}) + b_0 \mathbf{1} \right) \\
&= v_i \phi'(w_i \mathbf{x} + b_i \mathbf{1}) \odot \mathbf{x} \\
&= v_i \mathbf{x} \odot [\![w_i \mathbf{x} + b_i \mathbf{1} > 0]\!] \\
&= v_i \mathbf{x}_i \\
\frac{\partial \hat{\mathbf{f}}}{\partial v_i} &= \frac{\partial}{\partial v_i} \left( \sum_{i=1}^{H} v_i \phi(w_i \mathbf{x} + b_i \mathbf{1}) + b_0 \mathbf{1} \right) \\
&= \phi(w_i \mathbf{x} + b_i \mathbf{1}) \\
&= (w_i \mathbf{x} + b_i \mathbf{1}) \odot [\![w_i \mathbf{x} + b_i \mathbf{1} > 0]\!] \\
&= w_i \mathbf{x}_i + b_i \mathbf{1}_i \\
\frac{\partial \hat{\mathbf{f}}}{\partial b_i} &= \frac{\partial}{\partial b_i} \left( \sum_{i=1}^{H} v_i \phi(w_i \mathbf{x} + b_i \mathbf{1}) + b_0 \mathbf{1} \right) \\
&= v_i \phi'(w_i \mathbf{x} + b_i \mathbf{1})
\end{aligned}$$

$$
\begin{aligned}
&= v_i [\![ w_i \mathbf{x} + b_i \mathbf{1} > 0 ]\!] \\
&= v_i \mathbf{1}_i \\
\frac{\partial^2 \hat{\mathbf{f}}}{\partial b_0 \partial b_0} &= \frac{\partial}{\partial b_0} \mathbf{1} = \mathbf{0} \\
\frac{\partial^2 \hat{\mathbf{f}}}{\partial w_j \partial b_0} &= \frac{\partial}{\partial w_j} \mathbf{1} = \mathbf{0} \\
\frac{\partial^2 \hat{\mathbf{f}}}{\partial v_j \partial b_0} &= \frac{\partial}{\partial v_j} \mathbf{1} = \mathbf{0} \\
\frac{\partial^2 \hat{\mathbf{f}}}{\partial b_j \partial b_0} &= \frac{\partial}{\partial b_j} \mathbf{1} = \mathbf{0} \\
\frac{\partial^2 \hat{\mathbf{f}}}{\partial b_0 \partial w_i} &= \frac{\partial}{\partial b_0} v_i \mathbf{x}_i = \mathbf{0} \\
\frac{\partial^2 \hat{\mathbf{f}}}{\partial w_j \partial w_i} &= \frac{\partial}{\partial w_j} v_i \mathbf{x}_i = \delta_{ij} v_i \mathbf{x} \odot \delta(w_i \mathbf{x} + b_i \mathbf{1}) \odot \mathbf{x} \\
&= \delta_{ij} v_i \mathbf{x}_{i,=}^{\odot 2} \\
\frac{\partial^2 \hat{\mathbf{f}}}{\partial v_j \partial w_i} &= \frac{\partial}{\partial v_j} v_i \mathbf{x}_i = \delta_{ij} \mathbf{x}_i \\
\frac{\partial^2 \hat{\mathbf{f}}}{\partial b_j \partial w_i} &= \frac{\partial}{\partial b_j} v_i \mathbf{x}_i = v_i \mathbf{x} \odot \delta(w_i \mathbf{x} + b_i \mathbf{1}) \odot \mathbf{1} \\
&= \delta_{ij} v_i \mathbf{x}_{i,=} \\
\frac{\partial^2 \hat{\mathbf{f}}}{\partial b_0 \partial v_i} &= \frac{\partial}{\partial b_0} \left( w_i \mathbf{x}_i + b_i \mathbf{1}_i \right) = \mathbf{0} \\
\frac{\partial^2 \hat{\mathbf{f}}}{\partial w_j \partial v_i} &= \frac{\partial}{\partial w_j} \left( w_i \mathbf{x}_i + b_i \mathbf{1}_i \right) = \delta_{ij} \left[ \mathbf{x}_i + (w_i \mathbf{x} + b_i \mathbf{1}) \odot \delta \left( w_i \mathbf{x} + b_i \mathbf{1} \right) \odot \mathbf{x} \right] \\
&= \delta_{ij} \mathbf{x}_i \\
\frac{\partial^2 \hat{\mathbf{f}}}{\partial v_j \partial v_i} &= \frac{\partial}{\partial v_j} \left( w_i \mathbf{x}_i + b_i \mathbf{1}_i \right) = \mathbf{0} \\
\frac{\partial^2 \hat{\mathbf{f}}}{\partial b_j \partial v_i} &= \frac{\partial}{\partial b_j} \left( w_i \mathbf{x}_i + b_i \mathbf{1}_i \right) = \delta_{ij} \left[ \mathbf{1}_i + (w_i \mathbf{x} + b_i \mathbf{1}) \odot \delta \left( w_i \mathbf{x} + b_i \mathbf{1} \right) \odot \mathbf{1} \right] \\
&= \delta_{ij} \mathbf{1}_i \\
\frac{\partial^2 \hat{\mathbf{f}}}{\partial b_0 \partial b_i} &= \frac{\partial}{\partial b_0} v_i \mathbf{1}_i = \mathbf{0} \\
\frac{\partial^2 \hat{\mathbf{f}}}{\partial w_j \partial b_i} &= \frac{\partial}{\partial w_j} v_i \mathbf{1}_i = \delta_{ij} v_i \mathbf{1} \odot \delta \left( w_i \mathbf{x} + b_i \mathbf{1} \right) \odot \mathbf{1} \\
&= \delta_{ij} v_i \mathbf{1}_{i,=} \\
\frac{\partial^2 \hat{\mathbf{f}}}{\partial v_j \partial b_i} &= \frac{\partial}{\partial v_j} v_i \mathbf{1}_i = \delta_{ij} \mathbf{1}_i \\
\frac{\partial^2 \hat{\mathbf{f}}}{\partial b_j \partial b_i} &= \frac{\partial}{\partial b_j} v_i \mathbf{1}_i = \delta_{ij} v_i \mathbf{1} \odot \delta \left( w_i \mathbf{x} + b_i \mathbf{1} \right) \odot \mathbf{1} \\
&= \delta_{ij} \mathbf{1}_{i,=}
\end{aligned}
$$

Next, we calculate the first and second derivatives of $\ell$. Let $\varphi$ and $\psi$ denote arbitrary individual parameters.

$$
\begin{aligned}
\ell(\mathbf{x}) &= \frac{1}{2} \langle \hat{\boldsymbol{\epsilon}}, \hat{\boldsymbol{\epsilon}} \rangle \\
\frac{\partial \ell}{\partial \varphi} &= \langle \hat{\boldsymbol{\epsilon}}, \frac{\partial}{\partial \varphi} \hat{\boldsymbol{\epsilon}} \rangle = -\langle \hat{\boldsymbol{\epsilon}}, \frac{\partial}{\partial \varphi} \hat{\mathbf{f}} \rangle
\end{aligned}
$$

$$
\begin{aligned}
\frac{\partial \ell}{\partial b_0} &= -\langle \hat{\boldsymbol{\epsilon}}, \mathbf{1} \rangle \\
\frac{\partial \ell}{\partial w_i} &= -\langle \hat{\boldsymbol{\epsilon}}, v_i \mathbf{x}_i \rangle \\
\frac{\partial \ell}{\partial v_i} &= -\langle \hat{\boldsymbol{\epsilon}}, w_i \mathbf{x}_i + b_i \mathbf{1}_i \rangle \\
\frac{\partial \ell}{\partial b_i} &= -\langle \hat{\boldsymbol{\epsilon}}, v_i \mathbf{1}_i \rangle \\
\frac{\partial^2 \ell}{\partial \psi \partial \varphi} &= \langle \frac{\partial \hat{\boldsymbol{\epsilon}}}{\partial \psi}, \frac{\partial \hat{\boldsymbol{\epsilon}}}{\partial \varphi} \rangle + \langle \hat{\boldsymbol{\epsilon}}, \frac{\partial^2 \hat{\boldsymbol{\epsilon}}}{\partial \psi \partial \varphi} \rangle \\
&= \langle \frac{\partial \hat{\mathbf{f}}}{\partial \psi}, \frac{\partial \hat{\mathbf{f}}}{\partial \varphi} \rangle - \langle \hat{\boldsymbol{\epsilon}}, \frac{\partial^2 \hat{\mathbf{f}}}{\partial \psi \partial \varphi} \rangle \\
\frac{\partial^2 \ell}{\partial b_0 \partial b_0} &= \langle \frac{\partial \hat{\mathbf{f}}}{\partial b_0}, \frac{\partial \hat{\mathbf{f}}}{\partial b_0} \rangle - \langle \hat{\boldsymbol{\epsilon}}, \frac{\partial^2 \hat{\mathbf{f}}}{\partial b_0 \partial b_0} \rangle \\
&= \langle \mathbf{1}, \mathbf{1} \rangle \\
\frac{\partial^2 \ell}{\partial w_j \partial b_0} &= \langle \frac{\partial \hat{\mathbf{f}}}{\partial w_j}, \frac{\partial \hat{\mathbf{f}}}{\partial b_0} \rangle - \langle \hat{\boldsymbol{\epsilon}}, \frac{\partial^2 \hat{\mathbf{f}}}{\partial w_j \partial b_0} \rangle \\
&= \langle v_j \mathbf{x}_j, \mathbf{1} \rangle \\
\frac{\partial^2 \ell}{\partial v_j \partial b_0} &= \langle \frac{\partial \hat{\mathbf{f}}}{\partial v_j}, \frac{\partial \hat{\mathbf{f}}}{\partial b_0} \rangle - \langle \hat{\boldsymbol{\epsilon}}, \frac{\partial^2 \hat{\mathbf{f}}}{\partial v_j \partial b_0} \rangle \\
&= \langle w_j \mathbf{x}_j + b_j \mathbf{1}_j, \mathbf{1} \rangle \\
\frac{\partial^2 \ell}{\partial b_j \partial b_0} &= \langle \frac{\partial \hat{\mathbf{f}}}{\partial b_j}, \frac{\partial \hat{\mathbf{f}}}{\partial b_0} \rangle - \langle \hat{\boldsymbol{\epsilon}}, \frac{\partial^2 \hat{\mathbf{f}}}{\partial b_j \partial b_0} \rangle \\
&= \langle v_j \mathbf{1}_j, \mathbf{1} \rangle \\
\frac{\partial^2 \ell}{\partial b_0 \partial w_i} &= \langle \frac{\partial \hat{\mathbf{f}}}{\partial b_0}, \frac{\partial \hat{\mathbf{f}}}{\partial w_i} \rangle - \langle \hat{\boldsymbol{\epsilon}}, \frac{\partial^2 \hat{\mathbf{f}}}{\partial b_0 \partial w_i} \rangle \\
&= \langle \mathbf{1}, v_i \mathbf{x}_i \rangle \\
\frac{\partial^2 \ell}{\partial w_j \partial w_i} &= \langle \frac{\partial \hat{\mathbf{f}}}{\partial w_j}, \frac{\partial \hat{\mathbf{f}}}{\partial w_i} \rangle - \langle \hat{\boldsymbol{\epsilon}}, \frac{\partial^2 \hat{\mathbf{f}}}{\partial w_j \partial w_i} \rangle \\
&= \langle v_j \mathbf{x}_j, v_i \mathbf{x}_i \rangle - \langle \hat{\boldsymbol{\epsilon}}, \delta_{ij} v_i \mathbf{x}_{i,=}^{\odot 2} \rangle \\
\frac{\partial^2 \ell}{\partial v_j \partial w_i} &= \langle \frac{\partial \hat{\mathbf{f}}}{\partial v_j}, \frac{\partial \hat{\mathbf{f}}}{\partial w_i} \rangle - \langle \hat{\boldsymbol{\epsilon}}, \frac{\partial^2 \hat{\mathbf{f}}}{\partial v_j \partial w_i} \rangle \\
&= \langle w_j \mathbf{x}_j + b_j \mathbf{1}_j, v_i \mathbf{x}_i \rangle - \langle \hat{\boldsymbol{\epsilon}}, \delta_{ij} \mathbf{x}_i \rangle \\
\frac{\partial^2 \ell}{\partial b_j \partial w_i} &= \langle \frac{\partial \hat{\mathbf{f}}}{\partial b_j}, \frac{\partial \hat{\mathbf{f}}}{\partial w_i} \rangle - \langle \hat{\boldsymbol{\epsilon}}, \frac{\partial^2 \hat{\mathbf{f}}}{\partial b_j \partial w_i} \rangle \\
&= \langle v_j \mathbf{1}_j, v_i \mathbf{x}_i \rangle - \langle \hat{\boldsymbol{\epsilon}}, \delta_{ij} v_i \mathbf{x}_{i,=} \rangle \\
\frac{\partial^2 \ell}{\partial b_0 \partial v_i} &= \langle \frac{\partial \hat{\mathbf{f}}}{\partial b_0}, \frac{\partial \hat{\mathbf{f}}}{\partial v_i} \rangle - \langle \hat{\boldsymbol{\epsilon}}, \frac{\partial^2 \hat{\mathbf{f}}}{\partial b_0 \partial v_i} \rangle \\
&= \langle \mathbf{1}, w_i \mathbf{x}_i + b_i \mathbf{1}_i \rangle \\
\frac{\partial^2 \ell}{\partial w_j \partial v_i} &= \langle \frac{\partial \hat{\mathbf{f}}}{\partial w_j}, \frac{\partial \hat{\mathbf{f}}}{\partial v_i} \rangle - \langle \hat{\boldsymbol{\epsilon}}, \frac{\partial^2 \hat{\mathbf{f}}}{\partial w_j \partial v_i} \rangle \\
&= \langle v_j \mathbf{x}_j, w_i \mathbf{x}_i + b_i \mathbf{1}_i \rangle - \langle \hat{\boldsymbol{\epsilon}}, \delta_{ij} \mathbf{x}_i \rangle \\
\frac{\partial^2 \ell}{\partial v_j \partial v_i} &= \langle \frac{\partial \hat{\mathbf{f}}}{\partial v_j}, \frac{\partial \hat{\mathbf{f}}}{\partial v_i} \rangle - \langle \hat{\boldsymbol{\epsilon}}, \frac{\partial^2 \hat{\mathbf{f}}}{\partial v_j \partial v_i} \rangle \\
&= \langle w_j \mathbf{x}_j + b_j \mathbf{1}_j, w_i \mathbf{x}_i + b_i \mathbf{1}_i \rangle \\
\frac{\partial^2 \ell}{\partial b_j \partial v_i} &= \langle \frac{\partial \hat{\mathbf{f}}}{\partial b_j}, \frac{\partial \hat{\mathbf{f}}}{\partial v_i} \rangle - \langle \hat{\boldsymbol{\epsilon}}, \frac{\partial^2 \hat{\mathbf{f}}}{\partial b_j \partial v_i} \rangle
\end{aligned}
$$

$$
\begin{aligned}
&= \langle v_j \mathbf{1}_j, w_i \mathbf{x}_i + b_i \mathbf{1}_i \rangle - \langle \hat{\boldsymbol{\epsilon}}, \delta_{ij} \mathbf{1}_i \rangle \\
\frac{\partial^2 \ell}{\partial b_0 \partial b_i} &= \langle \frac{\partial \hat{\mathbf{f}}}{\partial b_0}, \frac{\partial \hat{\mathbf{f}}}{\partial b_i} \rangle - \langle \hat{\boldsymbol{\epsilon}}, \frac{\partial^2 \hat{\mathbf{f}}}{\partial b_0 \partial b_i} \rangle \\
&= \langle \mathbf{1}, v_i \mathbf{1}_i \rangle \\
\frac{\partial^2 \ell}{\partial w_j \partial b_i} &= \langle \frac{\partial \hat{\mathbf{f}}}{\partial w_j}, \frac{\partial \hat{\mathbf{f}}}{\partial b_i} \rangle - \langle \hat{\boldsymbol{\epsilon}}, \frac{\partial^2 \hat{\mathbf{f}}}{\partial w_j \partial b_i} \rangle \\
&= \langle v_j \mathbf{x}_j, v_i \mathbf{1}_i \rangle - \langle \hat{\boldsymbol{\epsilon}}, \delta_{ij} v_i \mathbf{1}_{i,=} \rangle \\
\frac{\partial^2 \ell}{\partial v_j \partial b_i} &= \langle \frac{\partial \hat{\mathbf{f}}}{\partial v_j}, \frac{\partial \hat{\mathbf{f}}}{\partial b_i} \rangle - \langle \hat{\boldsymbol{\epsilon}}, \frac{\partial^2 \hat{\mathbf{f}}}{\partial v_j \partial b_i} \rangle \\
&= \langle w_j \mathbf{x}_j + b_j \mathbf{1}_j, v_i \mathbf{1}_i \rangle - \langle \hat{\boldsymbol{\epsilon}}, \delta_{ij} \mathbf{1}_i \rangle \\
\frac{\partial^2 \ell}{\partial b_j \partial b_i} &= \langle \frac{\partial \hat{\mathbf{f}}}{\partial b_j}, \frac{\partial \hat{\mathbf{f}}}{\partial b_i} \rangle - \langle \hat{\boldsymbol{\epsilon}}, \frac{\partial^2 \hat{\mathbf{f}}}{\partial b_j \partial b_i} \rangle \\
&= \langle v_j \mathbf{1}_j, v_i \mathbf{1}_i \rangle - \langle \hat{\boldsymbol{\epsilon}}, \delta_{ij} \mathbf{1}_{i,=} \rangle
\end{aligned}
$$

Collecting all of the above together into the Hessian matrix, we have

$$
\mathbf{H}_\ell \triangleq \text{Hess}(\ell) \triangleq \begin{pmatrix}
\ddots & & & & \\
& \frac{\partial^2 \ell}{\partial w_j \partial w_i} & \frac{\partial^2 \ell}{\partial w_j \partial v_i} & \frac{\partial^2 \ell}{\partial w_j \partial b_i} & \cdots & \frac{\partial^2 \ell}{\partial w_j \partial b_0} \\
& \frac{\partial^2 \ell}{\partial v_j \partial w_i} & \frac{\partial^2 \ell}{\partial v_j \partial v_i} & \frac{\partial^2 \ell}{\partial v_j \partial b_i} & \cdots & \frac{\partial^2 \ell}{\partial v_j \partial b_0} \\
& \frac{\partial^2 \ell}{\partial b_j \partial w_i} & \frac{\partial^2 \ell}{\partial b_j \partial v_i} & \frac{\partial^2 \ell}{\partial b_j \partial b_i} & \cdots & \frac{\partial^2 \ell}{\partial b_j \partial b_0} \\
& \vdots & & & \ddots & \vdots \\
& \frac{\partial^2 \ell}{\partial b_0 \partial w_i} & \frac{\partial^2 \ell}{\partial b_0 \partial v_i} & \frac{\partial^2 \ell}{\partial b_0 \partial b_i} & \cdots & \frac{\partial^2 \ell}{\partial b_0^2}
\end{pmatrix}
$$

$$
= \begin{pmatrix}
\ddots & & & & \\
& \langle v_j \mathbf{x}_j, v_i \mathbf{x}_i \rangle - \langle \hat{\boldsymbol{\epsilon}}, \delta_{ij} v_i \mathbf{x}_{i,=}^{\odot 2} \rangle & \langle v_j \mathbf{x}_j, w_i \mathbf{x}_i + b_i \mathbf{1}_i \rangle - \langle \hat{\boldsymbol{\epsilon}}, \delta_{ij} \mathbf{x}_i \rangle & \langle v_j \mathbf{x}_j, v_i \mathbf{1}_i \rangle - \langle \hat{\boldsymbol{\epsilon}}, \delta_{ij} v_i \mathbf{1}_{i,=} \rangle & \cdots & \langle v_j \mathbf{x}_j, \mathbf{1} \rangle \\
& \langle w_j \mathbf{x}_j + b_j \mathbf{1}_j, v_i \mathbf{x}_i \rangle - \langle \hat{\boldsymbol{\epsilon}}, \delta_{ij} \mathbf{x}_i \rangle & \langle w_j \mathbf{x}_j + b_j \mathbf{1}_j, w_i \mathbf{x}_i + b_i \mathbf{1}_i \rangle & \langle w_j \mathbf{x}_j + b_j \mathbf{1}_j, v_i \mathbf{1}_i \rangle - \langle \hat{\boldsymbol{\epsilon}}, \delta_{ij} \mathbf{1}_i \rangle & \cdots & \langle w_j \mathbf{x}_j + b_j \mathbf{1}_j, \mathbf{1} \rangle \\
& \langle v_j \mathbf{1}_j, v_i \mathbf{x}_i \rangle - \langle \hat{\boldsymbol{\epsilon}}, \delta_{ij} v_i \mathbf{x}_{i,=} \rangle & \langle v_j \mathbf{1}_j, w_i \mathbf{x}_i + b_i \mathbf{1}_i \rangle - \langle \hat{\boldsymbol{\epsilon}}, \delta_{ij} \mathbf{1}_i \rangle & \langle v_j \mathbf{1}_j, v_i \mathbf{1}_i \rangle - \langle \hat{\boldsymbol{\epsilon}}, \delta_{ij} \mathbf{1}_{i,=} \rangle & \cdots & \langle v_j \mathbf{1}_j, \mathbf{1} \rangle \\
& \vdots & & & \ddots & \vdots \\
& \langle \mathbf{1}, v_i \mathbf{x}_i \rangle & \langle \mathbf{1}, w_i \mathbf{x}_i + b_i \mathbf{1}_i \rangle & \langle \mathbf{1}, v_i \mathbf{1}_i \rangle & \cdots & \langle \mathbf{1}, \mathbf{1} \rangle
\end{pmatrix}
$$

where $\mathbf{1}_{i,=}$ is the vector that selects datapoints $x_n$ that equal $\beta_i$, i.e. $\mathbf{1}_{(i,=),n} := [\![ w_i x_n + b_i = 0 ]\!]$, and $\mathbf{x}_{i,=} := \mathbf{x} \odot \mathbf{1}_{i,=}$. If we assume that $\beta_i \neq x_n$ for all $i$ and $n$ (which excludes a set of measure 0), we have

$$
= \begin{pmatrix}
\ddots & & & & \\
& \langle v_j \mathbf{x}_j, v_i \mathbf{x}_i \rangle & \langle v_j \mathbf{x}_j, w_i \mathbf{x}_i + b_i \mathbf{1}_i \rangle - \langle \hat{\boldsymbol{\epsilon}}, \delta_{ij} \mathbf{x}_i \rangle & \langle v_j \mathbf{x}_j, v_i \mathbf{1}_i \rangle & \cdots & \langle v_j \mathbf{x}_j, \mathbf{1} \rangle \\
& \langle w_j \mathbf{x}_j + b_j \mathbf{1}_j, v_i \mathbf{x}_i \rangle - \langle \hat{\boldsymbol{\epsilon}}, \delta_{ij} \mathbf{x}_i \rangle & \langle w_j \mathbf{x}_j + b_j \mathbf{1}_j, w_i \mathbf{x}_i + b_i \mathbf{1}_i \rangle & \langle w_j \mathbf{x}_j + b_j \mathbf{1}_j, v_i \mathbf{1}_i \rangle - \langle \hat{\boldsymbol{\epsilon}}, \delta_{ij} \mathbf{1}_i \rangle & \cdots & \langle w_j \mathbf{x}_j + b_j \mathbf{1}_j, \mathbf{1} \rangle \\
& \langle v_j \mathbf{1}_j, v_i \mathbf{x}_i \rangle & \langle v_j \mathbf{1}_j, w_i \mathbf{x}_i + b_i \mathbf{1}_i \rangle - \langle \hat{\boldsymbol{\epsilon}}, \delta_{ij} \mathbf{1}_i \rangle & \langle v_j \mathbf{1}_j, v_i \mathbf{1}_i \rangle & \cdots & \langle v_j \mathbf{1}_j, \mathbf{1} \rangle \\
& \vdots & & & \ddots & \vdots \\
& \langle \mathbf{1}, v_i \mathbf{x}_i \rangle & \langle \mathbf{1}, w_i \mathbf{x}_i + b_i \mathbf{1}_i \rangle & \langle \mathbf{1}, v_i \mathbf{1}_i \rangle & \cdots & \langle \mathbf{1}, \mathbf{1} \rangle
\end{pmatrix}
$$

At a critical point of $\ell$, we have $\langle \hat{\boldsymbol{\epsilon}}_i, \mathbf{x} \rangle = \langle \hat{\boldsymbol{\epsilon}}_i, \mathbf{1} \rangle = 0$, which yields further simplifications:

$$
= \begin{pmatrix}
\ddots & & & & \\
& \langle v_j \mathbf{x}_j, v_i \mathbf{x}_i \rangle & \langle v_j \mathbf{x}_j, w_i \mathbf{x}_i + b_i \mathbf{1}_i \rangle & \langle v_j \mathbf{x}_j, v_i \mathbf{1}_i \rangle & \cdots & \langle v_j \mathbf{x}_j, \mathbf{1} \rangle \\
& \langle w_j \mathbf{x}_j + b_j \mathbf{1}_j, v_i \mathbf{x}_i \rangle & \langle w_j \mathbf{x}_j + b_j \mathbf{1}_j, w_i \mathbf{x}_i + b_i \mathbf{1}_i \rangle & \langle w_j \mathbf{x}_j + b_j \mathbf{1}_j, v_i \mathbf{1}_i \rangle & \cdots & \langle w_j \mathbf{x}_j + b_j \mathbf{1}_j, \mathbf{1} \rangle \\
& \langle v_j \mathbf{1}_j, v_i \mathbf{x}_i \rangle & \langle v_j \mathbf{1}_j, w_i \mathbf{x}_i + b_i \mathbf{1}_i \rangle & \langle v_j \mathbf{1}_j, v_i \mathbf{1}_i \rangle & \cdots & \langle v_j \mathbf{1}_j, \mathbf{1} \rangle \\
& \vdots & & & \ddots & \vdots \\
& \langle \mathbf{1}, v_i \mathbf{x}_i \rangle & \langle \mathbf{1}, w_i \mathbf{x}_i + b_i \mathbf{1}_i \rangle & \langle \mathbf{1}, v_i \mathbf{1}_i \rangle & \cdots & \langle \mathbf{1}, \mathbf{1} \rangle
\end{pmatrix}
$$

as desired. □

**Corollary 2.** *Consider the setting of Theorem 5. There are seven cases where linear independence fails.*

*Proof.* Consider the following cases:

1. any neurons share activation patterns: the loss function doesn't care about what the function does between datapoints, so any change in one neuron that is "cancelled out" by other neuron(s) between it and the data leaves the loss unchanged
2. for every $i$, we have $w_i\mathbf{x}_i + b_i\mathbf{1}_i = c_1 v_i\mathbf{x}_i + c_2 v_i\mathbf{1}_i$ for $c_1 = \frac{w_i}{v_i}$ and $c_2 = \frac{b_i}{v_i}$: if $v_i$ is replaced by $\frac{v_i}{\alpha}$ and $w_i$ and $b_i$ are replaced by $\alpha w_i$ and $\alpha b_i$, respectively, $\hat{f}(x)$ (and thus the loss) is unchanged; thus there is a 1-dimensional hyperbolic submanifold of constant loss.
3. any neuron is active on the entire data: its bias is redundant with the global bias, leading to a 1-dimensional subspace of constant loss
4. any neuron is active on no data (i.e. $\mathbf{x}_i = \mathbf{1}_i = \mathbf{0}$): the breakpoint will be facing away from the data, so any change of its parameters that doesn't move the breakpoint into the data will have no effect on the loss ($v_i$ can change arbitrarily, $w_i$ and $b_i$ will have half-spaces of constant loss)
5. if for any $i$, $\mathbf{x}_i \propto \mathbf{1}_i$:
   - i.e. $w_i\mathbf{x}_i + b_i\mathbf{1}_i \equiv \alpha_i\mathbf{x}_i \propto v_i\mathbf{x}_i \propto v_i\mathbf{1}_i$
   - e.g. if $i$ has only one active datapoint or has multiple datapoints *all* with the same $x$-value
   - the line segment can freely rotate around its value at $x_n$
6. some of $(w_i, v_i, b_i)$ are 0
   - if $v_i = 0$, the delta-slope is 0, so the location of the breakpoint, and thus the values of $w_i$ and $b_i$ do not matter
   - if $w_i = 0$, the breakpoint is at infinity, so the values of $v_i$ and $b_i$ do not matter
   - if $b_i = 0$, changing $w_i$ will not move the breakpoint, so $w_i$ and $v_i$ can change so long as $w_i v_i$ remains constant
7. $\mathbf{1}_i + \mathbf{1}_j = \mathbf{1}$
   - the activation patterns of neuron $i$ and $j$ are mutually exclusive and collectively exhaustive
   - the $i$th and $j$th bias terms are redundant with the global bias $b_0$

Consider some features $b_1, b_2, b_3 \in \mathcal{B}$ and denote the activation patterns corresponding to these features by $\mathbf{1}_1, \mathbf{1}_2, \mathbf{1}_3$. Then, consider the case that $c_1 b_1 + c_2 b_2 \propto b_3$ for some constants $c_1$ and $c_2$. This can only happen if either (i) $\mathbf{1}_1 = \mathbf{1}_2 = \mathbf{1}_3$ which is handled by case (1), or (ii) $\mathbf{1}_1 + \mathbf{1}_2 = \mathbf{1}_3$. Because activation patterns are based on half-spaces, case (ii) can only happen if either $\mathbf{1}_3 = \mathbf{1}$, which is handled by case (7), or either $\mathbf{1}_1 = \mathbf{1}_3$ or $\mathbf{1}_2 = \mathbf{1}_3$, which is handled by case (1). Similar arguments apply for larger feature combinations, showing that the above list is exhaustive. □

**Corollary 3.** *Let $\theta^*$ be a critical point of $\ell(\cdot)$ and assume that its data partition is lonely, and at least one neuron is active on all data. Then the fraction of zero eigenvalues of the Hessian at $\theta^*$ is at least $1 - \frac{2N-3}{3H+1} \xrightarrow{H,N \gg 1} 1 - \frac{2}{3\mathcal{O}}$ where $\mathcal{O} \triangleq H/N$ is the overparametrization ratio.*

*Proof.* We begin by observing that each interior interval $[x_n, x_{n+1})$ for $n \in [N-1]$ contributes at most two linearly independent feature vectors: $\mathbf{x}_i$ and $\mathbf{1}_i$ for some $i$. Note that the presence of a neuron active on all data means that a neuron $j$ in the same interval, but with the opposite orientation will contribute features that are *not* linearly independent. Then, this gives us $2(N-1)$ vectors, then we have one additional vector for the global bias feature $\mathbf{1}$. In the best case, these vectors are all linearly independent, so we have

$$\begin{aligned}\operatorname{rank}\mathbf{H}_\ell^* &\equiv \dim\operatorname{span}\mathbf{H}_\ell^* \\ &\leq 2(N-1)+1 \\ &= 2N-1.\end{aligned}$$

The number of zero eigenvalues is then at least $(3H+1)-(2N-1)$ and so the fraction of zero eigenvalues is at least $1 - \frac{2N-1}{3H+1} \xrightarrow{H,N \gg 1} 1 - \frac{2}{3\mathcal{O}}$, as desired. □

### 5.8 Gradient Flow Dynamics in Spline Parameters

**Theorem 7.** *For a one hidden layer univariate ReLU network trained with gradient descent with respect to the neural network parameters $\theta_{NN} = \{(w_i, b_i, v_i)\}_{i=1}^H$, the gradient flow dynamics of the function space parameters $\theta_{BDSO} = \{(\beta_i, \mu_i)\}_{i=1}^H$ are governed by the following laws:*

$$\dot{\beta_i} = \frac{v_i(t)}{w_i(t)} [\underbrace{\langle \hat{\boldsymbol{\epsilon}}_i(t), \mathbf{1} \rangle}_{\text{net relevant residual}} + \beta_i(t) \underbrace{\langle \hat{\boldsymbol{\epsilon}}_i(t), \mathbf{x} \rangle}_{\text{correlation}}]$$

$$\dot{\mu_i} = w_i^2(t) \left[ - \left( 1 + \left( \frac{v_i(t)}{w_i(t)} \right)^2 \right) \langle \hat{\boldsymbol{\epsilon}}_i(t), \mathbf{x} \rangle + \beta_i(t) \langle \hat{\boldsymbol{\epsilon}}_i(t), \mathbf{1} \rangle \right]$$

*Proof.* Computing the time derivatives of the BDSO parameters and using the loss gradients of the loss with respect to the NN parameters gives us:

$$\begin{aligned}
\frac{\partial \ell(\theta_{\text{NN}})}{\partial w_i} &= v_i \langle \hat{\boldsymbol{\epsilon}}_i, \mathbf{x} \rangle \\
\frac{\partial \ell(\theta_{\text{NN}})}{\partial v_i} &= \langle \hat{\boldsymbol{\epsilon}}, \sigma(w_i \mathbf{x} + b_i \mathbf{1}) \rangle = \langle \hat{\boldsymbol{\epsilon}}_i, w_i \mathbf{x} + b_i \mathbf{1} \rangle = w_i \langle \hat{\boldsymbol{\epsilon}}_i, \mathbf{x} \rangle + b_i \langle \hat{\boldsymbol{\epsilon}}_i, \mathbf{1} \rangle \\
\frac{\partial \ell(\theta_{\text{NN}})}{\partial b_i} &= v_i \langle \hat{\boldsymbol{\epsilon}}_i, \mathbf{1} \rangle \\
\frac{\mathrm{d}\beta_i(t)}{\mathrm{d}t} &= \frac{\mathrm{d}}{\mathrm{d}t} \left( - \frac{b_i(t)}{w_i(t)} \right) \\
&= - \frac{w_i(t) \frac{\mathrm{d}b_i(t)}{\mathrm{d}t} - b_i(t) \frac{\mathrm{d}w_i(t)}{\mathrm{d}t}}{w_i(t)^2} \\
&= - \frac{w_i(t) (- \frac{\partial \ell(\theta_{NN})}{\partial b_i(t)}) - b_i(t) (- \frac{\partial \ell(\theta_{NN})}{\partial w_i(t)})}{w_i(t)^2} \\
&= \frac{w_i(t) \frac{\partial \ell(\theta_{NN})}{\partial b_i(t)} - b_i(t) \frac{\partial \ell(\theta_{NN})}{\partial w_i(t)}}{w_i(t)^2} \\
&= \frac{w_i(t) v_i(t) \langle \hat{\boldsymbol{\epsilon}}(t)_i(t), \mathbf{1} \rangle - b_i(t) v_i(t) \langle \hat{\boldsymbol{\epsilon}}(t)_i(t), \mathbf{x} \rangle}{w_i(t)^2} \\
&= \frac{v_i(t) \langle \hat{\boldsymbol{\epsilon}}(t)_i(t), w_i(t) \mathbf{1} - b_i(t) \mathbf{x} \rangle}{w_i(t)^2} \\
&= \frac{v_i(t)}{w_i(t)} \left\langle \hat{\boldsymbol{\epsilon}}(t)_i(t), \mathbf{1} - \frac{b_i(t)}{w_i(t)} \mathbf{x} \right\rangle \\
&= \frac{v_i(t)}{w_i(t)} \left\langle \underbrace{\hat{\boldsymbol{\epsilon}}(t)_i(t),}_{\text{relevant residuals}} \quad \mathbf{1} + \beta_i(t) \mathbf{x} \right\rangle \\
&= \frac{v_i(t)}{w_i(t)} [ \underbrace{\langle \hat{\boldsymbol{\epsilon}}(t)_i(t), \mathbf{1} \rangle}_{\text{net relevant residual}} + \beta_i(t) \underbrace{\langle \hat{\boldsymbol{\epsilon}}(t)_i(t), \mathbf{x} \rangle}_{\text{correlation}} ] \\
\frac{\mathrm{d}\mu_i(t)}{\mathrm{d}t} &= \frac{\mathrm{d}}{\mathrm{d}t} w_i v_i \\
&= \frac{\mathrm{d}w_i}{\mathrm{d}t} v_i + w_i \frac{\mathrm{d}v_i}{\mathrm{d}t} \\
&= - \frac{\partial \ell(\theta_{NN})}{\partial w_i} v_i - w_i \frac{\partial \ell(\theta_{NN})}{\partial v_i} \\
&= -v_i^2 \langle \hat{\boldsymbol{\epsilon}}_i, \mathbf{x} \rangle - w_i^2 \langle \hat{\boldsymbol{\epsilon}}_i, \mathbf{x} \rangle - w_i b_i \langle \hat{\boldsymbol{\epsilon}}_i, \mathbf{1} \rangle \\
&= -(v_i^2 + w_i^2) \langle \hat{\boldsymbol{\epsilon}}_i, \mathbf{x} \rangle - w_i b_i \langle \hat{\boldsymbol{\epsilon}}_i, \mathbf{1} \rangle \\
&= w_i^2(t) \left[ - \left( 1 + \left( \frac{v_i(t)}{w_i(t)} \right)^2 \right) \langle \hat{\boldsymbol{\epsilon}}_i(t), \mathbf{x} \rangle + \beta_i(t) \langle \hat{\boldsymbol{\epsilon}}_i(t), \mathbf{1} \rangle \right]
\end{aligned}$$

This completes the proof. □

### 5.9 Implicit Regularization

**Lemma 2.** *Consider the dynamics of gradient flow on $\ell(\cdot)$ started from $\theta_{NN,\alpha}(0) \triangleq (\alpha\mathbf{w}_0, \alpha\mathbf{b}_0, \mathbf{v}_0 = \mathbf{0})$, where $w_i \neq 0\ \forall i \in [H]$. In the limit $\alpha \to \infty$, $\boldsymbol{\beta}(t)$ does not change i.e. each breakpoint location is fixed. In this case, the $\theta_{NN}$ model reduces to a (kernel) linear regression:*

$$\hat{\mathbf{y}} = \boldsymbol{\Phi}(\mathbf{x}; \boldsymbol{\beta})\boldsymbol{\mu} \tag{9}$$

*where $\boldsymbol{\mu} \in \mathbb{R}^H$ are the regression weights and $\boldsymbol{\Phi}(\mathbf{x}; \boldsymbol{\beta}) \in \mathbb{R}^{N\times H}$ are the nonlinear features i.e. $\phi_{ni} \triangleq (x_n - \beta_i)_{s_i}$.*

*Proof.* First note that $\hat{f}(\cdot; \theta_{\text{NN}})$ is a 2-homogeneous model, i.e. $\hat{f}(\cdot; \lambda\theta_{\text{NN}}) = \lambda^2 \hat{f}(\cdot; \lambda\theta_{\text{NN}})$, and that $\hat{f}(\cdot; \theta_{\text{NN},\alpha}(0)) = 0$. Applying Theorem 2.2 of [7], we get that

$$\sup_{t\in[0,T]} \|\hat{f}(\mathbf{x}; \theta_{\text{NN},\alpha}(t)) - \bar{\hat{f}}(\mathbf{x}; \bar{\theta}_{\text{NN},\alpha}(t))\| = O(1/\alpha),$$

where the linearized model

$$\begin{aligned}
\bar{\hat{f}}(\mathbf{x}; \theta_{\text{NN},\alpha}) &\triangleq \hat{f}(\mathbf{x}; \theta_{\text{NN},\alpha}(0)) + \left\langle \frac{\partial \hat{f}(\mathbf{x}; \theta_{\text{NN},\alpha}(0))}{\partial \theta_{\text{NN},\alpha}}, \theta_{\text{NN},\alpha} - \theta_{\text{NN},\alpha}(0) \right\rangle \\
&= \left\langle \frac{\partial \hat{f}(\mathbf{x}; \theta_{\text{NN},\alpha}(0))}{\partial \theta_{\text{NN},\alpha}}, \theta_{\text{NN},\alpha} - \theta_{\text{NN},\alpha}(0) \right\rangle \\
&\triangleq \underset{N\times 3H}{\boldsymbol{\Phi}_{\text{NN}}} \underbrace{(\theta_{\text{NN},\alpha} - \theta_{\text{NN},\alpha}(0))}_{3H\times 1}
\end{aligned}$$

is the first order Taylor expansion of $\hat{f}(\mathbf{x}; \theta_{\text{NN},\alpha})$ about $\theta_{\text{NN},\alpha}(0)$, and $\bar{\theta}_{\text{NN},\alpha}(t)$ is the parameter trajectory of gradient flow according to the linearized model.

Next, observe that $\boldsymbol{\Phi}_{\text{NN}}$ is a matrix whose columns contain $3H$ feature vectors of the form $v_{0,i}\mathbf{x}_i$, $v_{0,i}\mathbf{1}_i$, and $w_{0,i}\mathbf{x}_i + b_{0,i}\mathbf{1}_i$ for each $i \in [H]$. Plugging these in and expanding the product, we have

$$\bar{\hat{f}}(\mathbf{x}; \theta_{\text{NN},\alpha}) = \sum_{i=1}^{H}(v_{0,i}\mathbf{x}_i)(w_i - w_{0,i}) + \sum_{i=1}^{H}(v_{0,i}\mathbf{1}_i)(b_i - b_{0,i}) + \sum_{i=1}^{H}(w_{0,i}\mathbf{x}_i + b_{0,i}\mathbf{1}_i)(v_i - v_{0,i})$$

Plugging in our $\theta_{\text{NN},\alpha}(0)$ we have $v_{0,i} = 0\ \forall i \in [H]$, giving

$$\begin{aligned}
&= \sum_{i=1}^{H}(w_{0,i}\mathbf{x}_i + b_{0,i}\mathbf{1}_i)v_i \\
&= \sum_{i=1}^{H} w_{0,i}v_i(\mathbf{x}_i - \beta_{0,i}\mathbf{1}_i) \\
&\triangleq \boldsymbol{\Phi}(\mathbf{x}, \boldsymbol{\beta})\boldsymbol{\mu}
\end{aligned}$$

□

**Theorem 6.** *Let $\boldsymbol{\mu}^*$ be the converged $\boldsymbol{\mu}$ parameter after gradient flow on the BDSO model Equation (7) starting from $\boldsymbol{\mu}_0 = \mathbf{0}$, with $\boldsymbol{\beta}$ held constant, and suppose that the model achieves $\tilde{\ell}(\theta_{BDSO}) = 0$. Then,*

$$\boldsymbol{\mu}^* = \arg\min_{\boldsymbol{\mu}} \|\boldsymbol{\mu}\|_2^2 \text{ s.t. } \boldsymbol{\Phi}\boldsymbol{\mu} = \mathbf{y}.$$

*Proof.* This proof follows the same proof strategy as the proof of Theorem 1 of [23]. First, we consider the dynamics of $\boldsymbol{\mu}(t)$ under gradient flow:

$$\dot{\boldsymbol{\mu}}(t) = -\frac{\partial \tilde{\ell}}{\partial \boldsymbol{\mu}} = -\left\langle \hat{\boldsymbol{\epsilon}}(t), \frac{\partial \hat{\boldsymbol{\epsilon}}(t)}{\partial \boldsymbol{\mu}} \right\rangle = \langle \hat{\boldsymbol{\epsilon}}(t), \boldsymbol{\Phi} \rangle .$$

Note that only the first term varies with time. Using this, we can integrate over time to get

$$\boldsymbol{\mu}^* = \boldsymbol{\mu}(\infty) = \boldsymbol{\mu}_0 + \int_0^\infty \langle \hat{\boldsymbol{\epsilon}}(t), \boldsymbol{\Phi}\rangle \,\mathrm{d}t = \left\langle \int_0^\infty \hat{\boldsymbol{\epsilon}}(t)\,\mathrm{d}t\,, \boldsymbol{\Phi} \right\rangle \triangleq \langle \mathbf{r}^*, \boldsymbol{\Phi}\rangle \tag{10}$$

where we define $\mathbf{r}^* \triangleq \int_0^\infty \hat{\boldsymbol{\epsilon}}(t)\,\mathrm{d}t$.

Next, consider the convex optimization problem

$$\arg\min_{\boldsymbol{\mu}} Q(\boldsymbol{\mu}) \text{ s.t. } \boldsymbol{\Phi}\boldsymbol{\mu} = \mathbf{y},$$

where $Q(\cdot)$ is some as-yet unspecified convex real-valued function. Then, the KKT conditions for this problem are

$$\begin{aligned} \boldsymbol{\Phi}\boldsymbol{\mu} &= \mathbf{y} \\ \exists \boldsymbol{\nu}^* \text{ s.t. } \nabla_{\boldsymbol{\mu}} Q(\boldsymbol{\mu}^*) &= \left\langle \boldsymbol{\nu}^*, \nabla_{\boldsymbol{\mu}} \hat{\mathbf{f}}(\boldsymbol{\mu}^*) \right\rangle = \langle \boldsymbol{\nu}^*, \boldsymbol{\Phi}\rangle\,. \end{aligned} \tag{11}$$

If we set $\boldsymbol{\nu}^* = \mathbf{r}^*$, then the right hand sides of Equations (10) and (11) are equal, implying that the left hand sides are also equal:

$$\nabla_{\boldsymbol{\mu}} Q(\boldsymbol{\mu}^*) = \boldsymbol{\mu}^*.$$

Integrating both sides with respect to each $\mu_i$, we have

$$Q(\boldsymbol{\mu}) = \sum_{i=1}^{H} \int_0^{\mu_i} \mu_i \,\mathrm{d}\mu_i = \sum_{i=1}^{H} \frac{1}{2}\mu_i^2 \propto \|\boldsymbol{\mu}\|_2^2$$

□

**Lemma 4.** *Let* $(\theta_{BDSO,H})_H \triangleq ((\mu_{i,H}, \beta_{i,H}, s_{i,H})_{i=1}^H)_H$ *be a sequence of width-*$H$ *parameter settings such that the breakpoints* $(\beta_{i,H})_{i=1}^H$ *are evenly spaced on the interval* $[a, b]$ *for every* $H$*,* $\lim_{H\to\infty} \mu_{i,H} = 0$ *for every* $i$*, and* $\theta_{BDSO,\infty} \triangleq \lim_{H\to\infty} \theta_{BDSO,H}$ *is well-defined. Let* $\hat{f}_\infty(\cdot; \theta_{BDSO,\infty}) \triangleq \lim_{H\to\infty} \hat{f}(\cdot; \theta_{BDSO,H})$*. Then,*

$$\hat{f}_\infty''(x; \theta_{BDSO,\infty}) = \lim_{H\to\infty} \frac{\mu_{i(x),H}}{\Delta\beta_{i(x),H}},$$

*where* $(\mu_{i(x),H})_H$ *is the sequence of delta-slopes corresponding to the unique sequence* $(\beta_{i(x),H})_H$ *of breakpoints that converge to* $x$*, and* $\Delta\beta_{i(x),H} \triangleq \beta_{p(i(x))+1,H} - \beta_{p(i(x)),H}$*.*

*Proof.* From Equation (1), we can derive

$$\hat{f}'(x; \theta_{\mathrm{BDSO},H}) = \sum_{i=1}^{H} \mu_i \begin{cases} [\![x > \beta_i]\!], & s_i = 1 \\ [\![x < \beta_i]\!], & s_i = -1 \end{cases} \tag{12}$$

$$\hat{f}''(x; \theta_{\mathrm{BDSO},H}) = \sum_{i=1}^{H} \mu_i s_i \delta(x - \beta_i). \tag{13}$$

For convenience, let $\beta_p$ be the $p^{\text{th}}$ smallest $\beta_i$:

$$= \sum_{p=1}^{H} \mu_p s_p \delta(x - \beta_p)$$

Multiplying each term by $\frac{\Delta\beta_p}{\Delta\beta_p} = 1$, we get

$$= \sum_{p=1}^{H} \Delta\beta_p \frac{s_p \mu_p}{\Delta\beta_p} \delta(x - \beta_p)$$

Next, note from Equations (12) and (13) that the term $s_p\mu_p$ is exactly the change in $\hat{f}'(\cdot; \theta_{\mathrm{BDSO},H})$ at $\beta_p$. Plugging this in, we get:

$$= \sum_{p=1}^{H} \Delta\beta_p \left[ \frac{\hat{f}'(x + \frac{\Delta\beta_p}{2}) - \hat{f}'(x - \frac{\Delta\beta_p}{2})}{\Delta\beta_p} \right] \delta(x - \beta_p)$$

Taking the $H \to \infty$ limit, we note that the overall expression has the form of a Riemann sum, and the bracketed expression is the definition of $\hat{f}''_\infty(\cdot; \theta_{\text{BDSO},\infty})$, and is well-defined because $(\mu_{i,H})_H \to 0$ by hypothesis:

$$\xrightarrow{H\to\infty} \int_a^b \hat{f}''_\infty(\beta, \theta_{\text{BDSO},\infty}) \delta(x - \beta) \, \mathrm{d}\beta$$
$$= \hat{f}''_\infty(x, \theta_{\text{BDSO},\infty})$$

□

**Lemma 5.** *Consider the setting of Theorem 6, and take the infinite width limit as in Lemma 4. Then,* $\lim_{H\to\infty}(\mu_{i,H}(t))_H = 0$ *for all* $i$ *and* $t$.

*Proof.* By hypothesis, $\mu_{i,H}(0) = 0$. Then, inspecting Equation (1), we can derive $\dot{\mu}_{i,H}(t) = \langle \hat{\boldsymbol{\epsilon}}(t), (\mathbf{x} - \beta_i \mathbf{1})_{s_i} \rangle$, and

$$\mu_{i,H}(t) = \left\langle \int_0^t \hat{\boldsymbol{\epsilon}}(t) \, \mathrm{d}t \, , (\mathbf{x} - \beta_i \mathbf{1})_{s_i} \right\rangle. \tag{14}$$

Thus, $\mu_{i,H}(\infty) = 0$ iff $\lim_{H\to\infty} \int_0^\infty \hat{\boldsymbol{\epsilon}}(t) \, \mathrm{d}t = \mathbf{0}$. Note that there is no dependence on $H$ in the r.h.s. of Equation (14), except through $\hat{\boldsymbol{\epsilon}}(t)$. Then, for any fixed $\hat{\boldsymbol{\epsilon}}(t)$, because $\hat{f}(\cdot; \theta_{\text{BDSO},H})$ is a sum over $H$ terms whose time derivative does not depend on $H$, we have that $\frac{\mathrm{d}}{\mathrm{d}t}\hat{f}(\cdot; \theta_{\text{BDSO},H})$ increases as $H \to \infty$. Therefore, $\frac{\mathrm{d}}{\mathrm{d}t}\hat{\boldsymbol{\epsilon}}(t)$ increases and $\hat{\boldsymbol{\epsilon}}(t)$ decays to $\mathbf{0}$ faster as $H \to \infty$. Then, $\lim_{H\to\infty} \hat{\boldsymbol{\epsilon}}(t) = \hat{\boldsymbol{\epsilon}}(0)[\![t = 0]\!]$, so $\int_0^\infty \hat{\boldsymbol{\epsilon}}(t) \, \mathrm{d}t = \mathbf{0}$, as desired. □

**Lemma 6.** *Consider the setting of Theorem 6. Then taking the infinite width limit as in Lemma 4,* $\lim_{H\to\infty} \hat{f}''(\cdot; \theta^*_{BDSO,H})$ *is a CPWL spline with knots at the datapoints* $x_n$.

*Proof.* Let $\hat{f}''_\infty(\cdot; \theta^*_{\text{BDSO}}) \triangleq \lim_{H\to\infty} \hat{f}''(\cdot; \theta^*_{\text{BDSO}})$. Then, we can build up a representation of $\hat{f}''_\infty(\cdot; \theta^*_{\text{BDSO}})$ as follows

$$\hat{f}''_\infty(\cdot; \theta^*_{\text{BDSO}}) = \lim_{H\to\infty} \hat{f}''(\cdot; \theta^*_{\text{BDSO}})$$
$$= \lim_{H\to\infty} \int_0^\infty \dot{\hat{f}}''(\cdot; \theta_{\text{BDSO}}(t)) \, \mathrm{d}t$$

Using Equation (1) and noting that with $\beta_i$ fixed by hypothesis, only $\mu_i$ varies with time, we can expand $\dot{\hat{f}}''(\cdot; \theta_{\text{BDSO}}(t))$:

$$= \lim_{H\to\infty} \int_0^\infty \sum_{i=1}^{H} \dot{\mu}_i(t) s_i \delta(x - \beta_i) \, \mathrm{d}t$$

Using the fact that $\dot{\mu}_i(t) \triangleq -\frac{\partial \tilde{\ell}}{\partial \mu_i}$, we can derive $\dot{\mu}_i(t)s_i = \langle \hat{\boldsymbol{\epsilon}}, (\mathbf{x} - \beta_i \mathbf{1})_{s_i} \rangle s_i = s_i \langle \hat{\boldsymbol{\epsilon}}_i, \mathbf{x}_i \rangle - s_i \beta_i \langle \hat{\boldsymbol{\epsilon}}, \mathbf{1}_i \rangle \triangleq r_{1,i}(t) + r_{x,i}(t)\beta_i$. Furthermore, note that the terms $r_{1,i}(t)$ and $r_{x,i}(t)$ only depend on the index $i$ through the mask $\mathbf{1}_i$ and the sign $s_i$; in other words, they are the same for all breakpoints with the same activation pattern. This pattern is wholly determined by the orientation $s_i$ and which data interval $(x_n, x_{n+1})$ that $\beta_i$ falls into, and therefore can only take on $2(n + 1)$ values. Note that the possible activation patterns can be indexed by $n$ and orientation $s$, and let $\varsigma_{n,s}$ denote the set of breakpoints with the activation pattern corresponding to $(n, s)$, we have

$$= \lim_{H\to\infty} \int_0^\infty \sum_{\substack{n=1 \\ s\in\{+,-\}}}^{N} \sum_{i\in\varsigma_{n,s}} (r_{1,i}(t) + r_{x,i}(t)\beta_i) \, \delta(x - \beta_i) \, \mathrm{d}t$$

Because $r_{1,i}(t) = r_{1,j}(t)$ if $i, j \in \varsigma_{n,s}$ (and likewise for $r_{x,i}(t)$), we can write

$$= \lim_{H\to\infty} \int_0^\infty \sum_{n,s} \sum_{i\in\varsigma_{n,s}} \left(r_{1,n,s}(t) + r_{x,n,s}(t)\beta_i\right) \delta(x - \beta_i)\,\mathrm{d}t$$

Equivalently, we can sort the elements of each $\varsigma_{n,s}$:

$$= \lim_{H\to\infty} \int_0^\infty \sum_{n,s} \sum_{p\in\varsigma_{n,s}} \left(r_{1,n,s}(t) + r_{x,n,s}(t)\beta_p\right) \delta(x - \beta_p)\,\mathrm{d}t$$

Next, letting $R^*_{1,n,s} \triangleq \int_0^\infty r_{1,n,s}(t)\,\mathrm{d}t$ and $R^*_{x,n,s} \triangleq \int_0^\infty r_{x,n,s}(t)\,\mathrm{d}t$, we can simplify:

$$= \lim_{H\to\infty} \sum_{n,s} \sum_{p\in\varsigma_{n,s}} \left(R^*_{1,n,s} + R^*_{x,n,s}\beta_p\right) \delta(x - \beta_p)$$

Then, we multiply by $\frac{\Delta\beta_p}{\Delta\beta_p}$, where $\Delta\beta_p \triangleq \beta_{p+1} - \beta_p$, and move the limit inside the first sum (which does not depend on $H$), so that the inner sum becomes a Riemann sum:

$$= \sum_{n,s} \lim_{H\to\infty} \sum_{p\in\varsigma_{n,s}} \Delta\beta_p \left(\frac{R^*_{1,n,s}}{\Delta\beta_p} + \frac{R^*_{x,n,s}}{\Delta\beta_p}\beta_p\right) \delta(x - \beta_p)$$

Next, consider the set $\{\beta_p | p \in \varsigma_{n,s}\}$: because these breakpoints all share the same activation pattern, they must all fall within the same interval $[x_n, x_{n+1})$ for some $n$, which we choose to be half-open by convention. Thus, we can evaluate the limit, so that the inner Riemann sum becomes an integral with limits $x_n$ and $x^-_{n+1}$, with the dummy integration variable $z$ replacing the $\beta_p$:

$$= \sum_{n,s} \int_{x_n}^{x^-_{n+1}} \left(\tilde{R}^*_{1,n,s} + \tilde{R}^*_{2,n,s} z\right) \delta(x - z)\,\mathrm{d}z\,,$$

where $\tilde{R}^*_{1,n,s} \triangleq \lim_{H\to\infty} \frac{R^*_{1,n,s}}{\Delta\beta_p}$ and $\tilde{R}^*_{2,n,s} \triangleq \lim_{H\to\infty} \frac{R^*_{x,n,s}}{\Delta\beta_p}$. Note that $R^*_{1,n,s} + R^*_{x,n,s}\beta_p$ is just $\mu_i(\infty)$ for the appropriate $i$, so these limits exist iff $\lim_{H\to\infty} \mu_i(\infty) = 0$. Applying Lemma 5 gives us the desired result.

Consider an arbitrary integral of the form $I(x) = \int_a^{b^-} f(z)\delta(x - z)\,\mathrm{d}z$. Then, $I(x) = f(x)$ if $x \in [a, b)$, and $I(x) = 0$ otherwise. Applying this, we get

$$= \sum_{n,s} \left(\tilde{R}^*_{1,n,s} + \tilde{R}^*_{2,n,s} x\right) [\![x \in [x_n, x_{n+1})]\!] \tag{15}$$

Thus, we have a function which is piecewise linear with knots at the datapoints $x_n$. □

**Corollary 4.** *Consider the setting of Theorem 6, with the additional assumption that the breakpoints are uniformly spaced, and let $H \to \infty$. Then the learned function $\hat{f}_\infty(x; \boldsymbol{\mu}^*, \boldsymbol{\beta})$ is the global minimizer of*

$$\inf_f \int_{-\infty}^{\infty} f''(x)^2\,\mathrm{d}x \;\text{ s.t. } y_n = f(x_n)\;\forall n \in [N],$$

*As such, $\hat{f}(x; \boldsymbol{\mu}^*, \boldsymbol{\beta}^*)$ is a natural cubic smoothing spline with $N$ degrees of freedom [3].*

*Proof.* Take the limit as $H \to \infty$ of each step in the proof of Theorem 6 that requires it. This amounts to making the replacements

$$\begin{aligned} \boldsymbol{\mu} \in \mathcal{H}_H \equiv \mathbb{R}^H &\mapsto \mu(\cdot) \in \mathcal{H} \equiv \mathcal{L}_2([a,b]) \\ \langle\cdot,\cdot\rangle_{\mathcal{H}_H} &\mapsto \langle\cdot,\cdot\rangle_{\mathcal{H}} \\ \frac{\partial}{\partial\boldsymbol{\mu}} &\mapsto \frac{\delta}{\delta\boldsymbol{\mu}(\cdot)} \end{aligned}$$

$$\min \mapsto \inf$$
$$\|\boldsymbol{\mu}\|_2^2 \mapsto \int_{-\infty}^{\infty} \hat{f}''(x)^2 \, \mathrm{d}x$$

For the last replacement, use Lemma 4 to see that minimizing

$$\begin{aligned}
\|\boldsymbol{\mu}\|_2^2 &= \sum_{i=1}^{H} \mu_i^2 \\
&\propto \frac{1}{\Delta\beta} \sum_{i=1}^{H} \mu_i^2 \\
&= \frac{1}{\Delta\beta} \sum_{i=1}^{H} (s_i \mu_i)^2 \\
&= \sum_{i=1}^{H} \frac{(s_i \mu_i)^2}{\Delta\beta_i} \\
&\propto \sum_{i=1}^{H} \left( \frac{s_i \mu_i}{\Delta\beta_i} \right)^2
\end{aligned}$$

is equivalent to minimizing $\int_{-\infty}^{\infty} \hat{f}''(x)^2 \, \mathrm{d}x$ in the infinite width limit.

□